%% file: main.tex
\documentclass[journal]{IEEEtran}
\usepackage{amsmath,amsfonts}
\usepackage{algorithmic}
\usepackage{algorithm}
\usepackage{array}
\usepackage[caption=false,font=normalsize,labelfont=sf,textfont=sf]{subfig}
\usepackage{textcomp}
\usepackage{stfloats}
\usepackage{url}
\usepackage{verbatim}
\usepackage{graphicx}
\usepackage{tabularx}
\usepackage{cite}
\input{packages}
\input{tab/templates}
\begin{document}
\title{ELP-Adapters: Parameter Efficient Adapter Tuning for Various Speech Processing Tasks}
\author{
Nakamasa~Inoue,
Shinta Otake, Takumi Hirose, Masanari Ohi, Rei Kawakami,\\
Tokyo Institute of Technology
}

\markboth{}{}

\maketitle
\input{sec/0_abstract}
\input{sec/1_introduction}
\input{sec/2_conventional_methods}

\input{sec/3_proposed_method}
\input{sec/4_asr}

\input{sec/5_asv}
\input{sec/6_ser}
\input{sec/7_ic}
\input{sec/9_analysis}

\input{sec/10_conclusion}
\bibliography{main.bib}
\bibliographystyle{IEEEtran}
\end{document}

%% file: packages.tex
\usepackage{tcolorbox}
\usepackage{amsmath,amssymb}
\usepackage{algorithmic}
\usepackage{algorithm}
\usepackage{bm}
\usepackage{booktabs}
\usepackage{comment}
\usepackage{multirow}
\usepackage{framed}

\usepackage{url}
\makeatletter
\newcommand{\figcaption}[1]{\def\@captype{figure}\caption{#1}}
\newcommand{\tblcaption}[1]{\def\@captype{table}\caption{#1}}
\makeatother

\newcount\K
\def\bla#1{
\K=0 \loop\ifnum\K<#1
{\textcolor[gray]{0.9}{{\it bla bla bla bla bla bla bla bla bla bla bla bla bla bla bla}}}
\advance\K by1\repeat
}

\newcommand{\todo}[1]{
\ifx#10
\textcolor{red}{$0.00_{\pm 0.00}$}
\else
\textcolor{red}{#1}
\fi
}

\usepackage{cleveref}
\crefname{figure}{Fig.}{Figs.}
\Crefname{figure}{Fig.}{Figs.}
\crefname{table}{Table.}{Tables.}
\Crefname{table}{Table.}{Tables.}
\crefname{equation}{Eq.}{Eqs.}
\Crefname{equation}{Eq.}{Eqs.}
\crefname{section}{Section}{Sections}
\Crefname{section}{Section}{Sections}

%% file: tab/templates.tex
\newcommand{\tabresulttemplate}[7]{
\begin{table*}[t]
\centering
{
\setlength{\tabcolsep}{8.7pt}
#1
\begingroup
\renewcommand{\arraystretch}{1.1}
\begin{tabular}{l|c|ccccc|c}
\toprule
\hspace{-5pt}Fine-tuning method & \# Params & Wav2vec2.0 & HuBERT &  ContentVec & WavLM & WavLM+ & Average\\
\midrule
\hspace{-5pt}Weight tuning & $0.037\mathrm{M}$ & #2 \\
\hspace{-5pt}Prefix tuning & $14.81\mathrm{M}$ & #3\\
\hspace{-5pt}LoRA tuning & $9.47\mathrm{M}$ & #4 \\
\hspace{-5pt}Efficient adapter tuning & $9.54\mathrm{M}$ & #5 \\
\hspace{-5pt}ELP-adapter tuning (ours) & $9.52\mathrm{M}$ & #6\\
\midrule
\hspace{-5pt}Full fine-tuning & $94.70\mathrm{M}$ & #7\\ 
\bottomrule
\end{tabular}
\endgroup
}
}

\newcommand{\tabalbationtemplate}[6]{
\vspace{12pt}
\centering
{
\setlength{\tabcolsep}{10pt}
#1
\begingroup
\renewcommand{\arraystretch}{1.1}
\begin{tabular}{l|c|ccccc|c}
\toprule
\hspace{-5pt}Adapters & \# Params & Wav2vec2.0 & HuBERT &  ContentVec & WavLM & WavLM+ & Average\\
\midrule
\hspace{-5pt}P-adapter tuning  & $0.004\mathrm{M}$ & #3\\
\hspace{-5pt}E-adapter tuning & $4.79\mathrm{M}$ & #4 \\
\hspace{-5pt}L-adapter tuning & $4.77\mathrm{M}$ & #5 \\
\hspace{-5pt}EL-adapter  tuning& $9.13\mathrm{M}$ & #6 \\
\hspace{-5pt}ELP-adapter tuning & $9.52\mathrm{M}$ & #2\\
\bottomrule
\end{tabular}
\endgroup
}
\end{table*}
}

%% file: sec/0_abstract.tex
\begin{abstract}
Self-supervised learning has emerged as a key approach for learning generic representations from speech data. Despite promising results in downstream tasks such as speech recognition, speaker verification, and emotion recognition, a significant number of parameters is required, which makes fine-tuning for each task memory-inefficient.
To address this limitation, we introduce ELP-adapter tuning, a novel method for parameter-efficient fine-tuning using three types of adapter, namely
encoder adapters (E-adapters), layer adapters (L-adapters), and a prompt adapter (P-adapter).
The E-adapters are integrated into transformer-based encoder layers and help to learn fine-grained speech representations that are effective for speech recognition. The L-adapters create paths from each encoder layer to the downstream head and help to extract non-linguistic features from lower encoder layers that are effective for speaker verification and emotion recognition. The P-adapter appends pseudo features to CNN features to further improve effectiveness and efficiency.
With these adapters, models can be quickly adapted to various speech processing tasks.
Our evaluation across four downstream tasks using five backbone models demonstrated the effectiveness of the proposed method.
With the WavLM backbone, its performance was comparable to or better than that of full fine-tuning on all tasks while requiring 90\% fewer learnable parameters. 
\end{abstract}


%% file: sec/1_introduction.tex
\section{Introduction}
In the field of audio and speech processing, self-supervised learning using large-scale unlabeled datasets has become a leading approach for extracting generic representations from speech data~\cite{schneider2020wav2vec, baevski2020wav2vec2, wang2021unispeech, hsu2021hubert, chen2022unispeech-sat, chen2022wavlm, qian2022contentvec}. The main idea of this approach is to leverage the inherent structures and patterns within the speech data to train models via representation learning loss such as contrastive loss~\cite{chen2020simclr, jiang2021speechsimclr, huh2020augmentation, inoue2020semi}. This significantly reduces the need for manually labeled data, making model training more scalable and efficient.
Examples of models trained by self-supervised learning, which we refer to as self-supervised models,
include wav2vec~\cite{schneider2020wav2vec, baevski2020wav2vec2}, HuBERT~\cite{hsu2021hubert}, and WavLM~\cite{chen2022wavlm}. These models have demonstrated the ability to extract task-independent features with transformer-based architectures.

In recent years, the range of speech processing tasks that can be covered by self-supervised models has been steadily expanding beyond automatic speech recognition.
For example, a number of studies have proposed methods that utilize speech embeddings extracted from self-supervised models for discriminative tasks such as speaker verification\cite{finetune-sv, fan21wav2vecsv, lee22wav2vecsv, peng2023improving} and speech emotion recognition~\cite{finetune-er, pepino21wav2vecer}.
Some pioneering studies have demonstrated the effectiveness of self-supervised models in addressing more complex and generative tasks.
For example, spoken question answering is an important line of research focused on developing models capable of understanding and responding to questions posed in natural spoken language, where recent studies leverage self-supervised models~\cite{sqa1, sqa2, sqa3, sqa4, sqa5, sqa6}.
It has also been demonstrated that self-supervised models can perform voice conversion effectively and efficiently by integrating a decoder and a vocoder~\cite{vc0, vc1, vc2, vc3, vc4}.
These studies highlight the potential of self-supervised models across various speech tasks.

To apply self-supervised models to downstream tasks, fine-tuning on task-specific labeled datasets is often required.
This process enables the models to adapt and specialize in specific tasks, leading to excellent results not only in speech recognition but also in various speech tasks.
However, one limitation is the substantial number of parameters involved. When fine-tuning is conducted for each downstream task, multiple models must be stored, one for each task. This can lead to storage inefficiencies in real-world application settings, such as when each user wants to fine-tune the model with their private data and task.

A parameter-efficient method for adapting self-supervised models to various downstream tasks is thus desirable.
Learning task-specific downstream head modules, such as a linear classification head, with frozen self-supervised models is an efficient solution; however, it often degrades the final performance compared to that obtained by fine-tuning all parameters because the optimal features can differ substantially depending on each task. For instance, linguistic features that include phoneme information are crucial for speech recognition, whereas non-linguistic features are crucial for speaker verification.

Recently, learning with adapter modules that can be inserted into the intermediate encoder layers of a frozen model has emerged as a promising approach for parameter-efficient fine-tuning.
The first adapter tuning method~\cite{nlp-adapter} was proposed for BERT \cite{bert} in the field of natural language processing,
where two adapter modules are inserted into each encoder layer of BERT.
Each adapter module consists of two linear layers with an activation between them and a skip connection. This approach requires fewer parameters (the frozen parameters are shared among all downstream tasks) without degrading accuracy.
A number of follow-up studies have used adapters for various natural language processing tasks \cite{lin2020exploring, Guo2021AdaptiveAdapters, Zhang2023PoE}.

For speech recognition, Kannan~{et al.}~\cite{rnnt-adapter} integrated adapter modules into recurrent neural network transducers.
Hou~{et al.}~\cite{hou21meta, hou22adapters}
proposed the use of adapters for cross-lingual speech adaptation.
Winata~{et al.}~\cite{ctcattn-adapter} proposed the adapt-and-adjust framework, which uses adapter modules for multilingual speech recognition based on hybrid connectionist temporal classification (CTC)-attention networks.
Qian~{et al.}~\cite{Qian2022LayerWiseFastAdaptation} proposed gated and multi-basis adapters for multi-accent speech recognition.
The effectiveness of adapter tuning has also been demonstrated in other speech processing tasks such as speech translation~\cite{sl-adapter}.

Some recent studies have explored the application of adapter tuning to self-supervised models.
Thomas~{et al.}~\cite{speech-adapter} introduced adapter modules into wav2vec2.0 for speech recognition.
Chen~{et al.}~\cite{Chen2022EfficientTuning} compared the adapter modules with other efficient fine-tuning methods such as low-rank adaptation (LoRA)~\cite{hu22lora}.
It is also reported that various acoustic and linguistic features tend to
be encoded in different layers in wav2vec2.0~\cite{analysis1, analysis2}.
These studies inspired us to develop an adapter tuning method for not only speech recognition but also various other speech processing tasks.

In this work, we propose ELP-adapter tuning, a parameter-efficient fine-tuning method that utilizes three types of adapter, namely encoder adapters (E-adapters), layer adapters (L-adapters), and a prompt adapter (P-adapter). Each adapter is a small learnable module that has a distinct role in enhancing performance in downstream tasks.
Given a frozen self-supervised model that consists of multiple encoder layers, the E-adapters are integrated into the transformer-based encoder layers. They help to extract fine-grained linguistic representations and improve speech recognition performance. The L-adapters create paths from each encoder layer to the downstream head. This improves the performance of tasks such as emotion recognition and speaker verification, as features extracted from intermediate encoder layers often help to capture non-linguistic features. The P-adapter appends learnable embeddings that are used as auxiliary inputs to the transformer-based encoders. This further enhances learning effectiveness and efficiency.

In experiments, we applied ELP-adapter tuning to four downstream tasks, namely automatic speech recognition (ASR), automatic speaker verification (ASV), speech emotion recognition (SER), and speech intent classification (SIC).
With the WavLM backbone, our method achieved performance comparable to or even better than that of full fine-tuning while using 90\% fewer learnable parameters. Further, we visualized the weight coefficients for each layer to explain the improvement obtained with our method.

This paper is an extended version of our previously published paper~\cite{otake2023parameter} at ICASSP 2023. Compared to the previous version, we have made the following significant extensions:
\begin{enumerate}
\item We introduced a P-adapter that can be utilized in conjunction with the previously proposed E-adapters and L-adapters.
\item We demonstrated the effectiveness of ELP-adapter tuning across multiple self-supervised models. Specifically, we expanded our evaluations to include wav2vec2.0~\cite{baevski2020wav2vec2}, HuBERT~\cite{hsu2021hubert}, ContentVec~\cite{qian2022contentvec}, and WavLM+~\cite{chen2022wavlm}.
\item We thoroughly conducted experimental evaluation with multiple conventional fine-tuning methods including
weight tuning~\cite{chen2022wavlm}, LoRA tuning~\cite{hu22lora}, Prefix tuning~\cite{li21prefixtuning}, and Efficient adapter tuning~\cite{speech-adapter}.
\end{enumerate}

The rest of this paper is organized as follows.
Section~\ref{sec:conventional} reviews conventional self-supervised models and fine-tuning methods.
Section~\ref{sec:method} introduces ELP-adapter tuning for parameter-efficient fine-tuning.
Sections~\ref{sec:ASR}-\ref{sec:SIC}
respectively present experiments on ASR, ASV, SER, and SIC tasks.
Section~\ref{sec:analysis} provides detailed analysis on layer weights and adapter configurations.
Finally, Section~\ref{sec:conclusion} concludes this paper and discusses future research directions.

%% file: sec/2_conventional_methods.tex
\section{Conventional methods}
\label{sec:conventional}
\subsection{Self-supervised models}

The goal of self-supervised learning is to learn features from unlabeled data by leveraging the intrinsic structure of the data itself. This approach involves creating tasks where the input data serve as their own supervision data. 
Below, we review five self-supervised models for speech signal processing that we use as the backbones in our experiments.

\subsubsection{wav2vec2.0~\cite{baevski2020wav2vec2}}
This model consists of a convolutional neural network (CNN) encoder followed by multiple transformer encoders.
The CNN encoder extracts low-level features from raw waveform inputs via a sequence of several blocks, each with a temporal convolution layer, layer normalization~\cite{ba16layernorm}, and a Gaussian error linear unit (GELU)~\cite{Hendrycks16GELU} activation function.
The transformer encoders apply attention modules to the extracted features.
We employ the wav2vec2.0 base model trained on the Librispeech~\cite{panayotov2015librispeech} corpus, which contains 960 hours of speech with contrastive loss and diversity loss.
The number of parameters is 95.04M.

\subsubsection{HuBERT~\cite{hsu2021hubert}}
This model aims to discover hidden acoustic units to provide frame-level targets in self-supervised learning using masked prediction.
The architecture consists of a CNN encoder and transformer encoders, similar to wav2vec2.0.
We employ the HuBERT base model, which is trained on the Librispeech corpus with masked prediction loss using the acoustic unit discovery module.
The number of parameters is 94.68M.

\subsubsection{ContentVec~\cite{qian2022contentvec}}
This model aims to disentangle speaker variations during self-supervised learning by incorporating three disentanglement mechanisms into HuBERT, namely disentanglement in teachers, students, and speaker conditioning.
The architecture is the same as that of the HuBERT base model.
We employ the ContentVec model trained on Librispeech.

\subsubsection{WavLM~\cite{chen2022wavlm}}
This model is a self-supervised model for addressing various downstream speech tasks.
The architecture consists of a CNN encoder and transformer-based encoders using gated relative position bias~\cite{chi2022xlm}.
We employ two models, namely WavLM Base and WavLM Base+.
The former model is trained on Librispeech.
The latter model, which we refer to as WavLM+, is trained on a union set of Librispeech, GigaSpeech~\cite{chen21gigaspeech}, and VoxPopuli~\cite{wang21voxpopuli}, which contains a total of 96k hours of audio data.
The number of parameters for each model is 94.70M.

\subsection{Fine-tuning methods}

Given a self-supervised model, the goal of fine-tuning is to adjust the model parameters for a specific downstream task, typically using a relatively small amount of labeled data and a task-specific loss function.
Assuming that self-supervised models share a common architecture, which consists of a CNN encoder followed by multiple transformer-based encoders as shown in \cref{fig:conventional}(a), below we provide details on five fine-tuning methods that we use as baselines in our experiments.

\label{sec:fine-tuning_methods}
\subsubsection{Full fine-tuning}
This method updates all model parameters for each downstream task.
Typically, a small downstream head such as a linear head or a multi-layer perceptron (MLP) with few layers is added to the self-supervised model to apply a task-specific loss function
such as CTC loss for ASR~\cite{graves2006ctc} and cross entropy loss for SIC.
In general, full fine-tuning is less parameter-efficient, but it often achieves high performance on downstream tasks.

\subsubsection{Weight tuning}
This method utilizes the weighted sum of features extracted from the encoder layers, where the weight coefficients are learnable and the other parameters of the self-supervised model are frozen as shown in \cref{fig:conventional}(b).
More specifically, it is formulated as
\begin{align}
\label{weight_tuning}
\bar{X} = \sum_{l=1}^{L} w_{l} X_{l},
\end{align}
where $X_{l} \in \mathbb{R}^{n \times d}$ is the output of the $l$-th encoder layer given as a sequence of $d$-dimensional vectors of length $n \in \mathbb{N}$,
$w_{l}\hspace{-3pt}~\in~\hspace{-3pt}\mathbb{R}$ is a learnable weight, and $L \in \mathbb{N}$ is the number of encoder layers.
When applying weight tuning to downstream tasks, a learnable downstream head that takes $\bar{X} \in \mathbb{R}^{n \times d}$ as the input is added to the frozen self-supervised model.
As discussed in~\cite{chen2022wavlm}, this method is significantly more parameter-efficient than full fine-tuning because most parameters are frozen and shared among all downstream tasks. However, performance on downstream tasks is often degraded.
\input{fig/fig_conventional}

\subsubsection{LoRA tuning~\cite{hu22lora}}
This method injects rank decomposition matrices into a frozen self-supervised model.
When applying LoRA tuning to self-attention modules, the key, value, and query matrices are computed with injected low-rank matrices as shown in \cref{fig:conventional}(c).
More specifically, given an input sequence $X \in \mathbb{R}^{n \times d}$, the self-attention module of LoRA tuning is given as
\begin{align}
f_{\text{attn}}(X)
&= \mathrm{softmax}\left(\frac{Q (K)^{\top}}{\sqrt{d}} \right) V,
\end{align}
where $K$, $V$, and $Q$ are the key, value, and query matrices, respectively, given by
\begin{align}
K &= X (W_{k} + A_{k}B_{k}),\\
V &= X (W_{v} + A_{v}B_{v}),\\
Q &= X (W_{q} + A_{q}B_{q}).
\end{align}
Here,
$W_{k}, W_{v}, W_{q} \in \mathbb{R}^{d \times d'}$ are pre-trained frozen weights,
$A_{k}, A_{v}, A_{q} \in \mathbb{R}^{d \times r}$ and 
$B_{k}, B_{v}, B_{q} \in \mathbb{R}^{r \times d^{\prime}}$ are learnable low-rank matrices. The rank $r$ is chosen such that $r \ll \min(d, d^{\prime})$.
We apply LoRA tuning to all self-attention modules and the fully connected layers after each self-attention module with $r = 128$.

\subsubsection{Prefix tuning~\cite{li21prefixtuning}}
This method prepends pseudo tokens to each encoder layer by concatenating new learnable embeddings to the key and value matrices of each self-attention module as shown in \cref{fig:conventional}(d).
Specifically, the key, value, and query matrices to compute self-attention are given by
\begin{align}
K &= [P_{k}; XW_{k}] \in \mathbb{R}^{(n+m) \times d^{\prime}},\\
V &= [P_{v}; XW_{v}] \in \mathbb{R}^{(n+m) \times d^{\prime}},\\
Q &= XW_{q}  \in \mathbb{R}^{n \times d^{\prime}}.
\end{align}
Here, $W_{k}, W_{v}, W_{q} \in \mathbb{R}^{d \times d^{\prime}}$ are pre-trained frozen weights,
$P_{k}, P_{v} \in \mathbb{R}^{m \times d^{\prime}}$ are newly added learnable matrices, and $[~;~]$ indicates the concatenation operation.
We apply prefix tuning to all self-attention modules with $m=5$.

\subsubsection{Efficient adapter tuning~\cite{speech-adapter}}
In the field of natural language processing, Houlsby~{\it et al.}~\cite{nlp-adapter} proposed efficient adapter modules for transformers.
This was applied to wav2vec2.0 by Thomas~{\it et al.}~\cite{speech-adapter} for speech recognition.
We refer to this method as efficient adapter tuning.
Let $X_{0} \in \mathbb{R}^{n \times d}$ be the
output of the CNN encoder, where $n$ is the length, which depends on the time length of the audio input, and $d$ is the dimension of feature vectors.
Under the assumption that the output $X_{l} \in \mathbb{R}^{n \times d}$ of the $l$-th encoder layer is given by
\begin{align}
\label{eq:enc2}
Z_{l} &= f_{\text{norm}}(f_{\text{mhsa}}^{(l)}(X_{l-1}) + X_{l-1}), \\
\label{eq:enc1}
X_{l} &= f_{\text{norm}}(f_{\text{ffn}}^{(l)}(Z_{l}) + Z_{l}), 
\end{align}
where $f_{\text{ffn}}^{(l)}$ is a feedforward network, $f_{\text{mhsa}}^{(l)}$ is a multi-head self-attention (MHSA) module, and $f_{\text{norm}}$ is a normalization function,
efficient adapter tuning inserts two learnable adapters $g^{(l)}_{1}$ and $g^{(l)}_{2}$ as follows:
\begin{align}
\label{eq:enca2}
\hat{Z}_{l} &= f_{\text{norm}}(g^{(l)}_{1}(f_{\text{mhsa}}^{(l)}(\hat{X}_{l-1})) + \hat{X}_{l-1}), \\
\label{eq:enca1}
\hat{X}_{l} &= f_{\text{norm}}(g^{(l)}_{2}(f_{\text{ffn}}^{(l)}(\hat{Z}_{l})) + \hat{Z}_{l}),
\end{align}
where $\hat~$ indicates adapted output features. Here, each adapter $g^{(l)}_{i}: \mathbb{R}^{n \times d} \to \mathbb{R}^{n \times d}~(i = 1, 2)$ is given by
\begin{align}
\label{nlpadapter}
g^{(l)}_{i} (X) = f_{\text{norm}}(f_{\text{fc2}}^{(l)}( \sigma( f_{\text{fc1}}^{(l)}(X)))) + X
\end{align}
where $f_{\text{fc1}}^{(l)}$ and $f_{\text{fc2}}^{(l)}$ are learnable fully connected layers and $\sigma$ is an activation function.
As shown in \cref{fig:conventional}(e), LayerNorm and GELU activation function are used for $f_{\text{norm}}$ and $\sigma$, respectively. A downstream head is also trained with the adapter modules.

\input{fig/fig_overview}

%% file: fig/fig_conventional.tex
\begin{figure*}
\centering
\includegraphics[width=\linewidth]{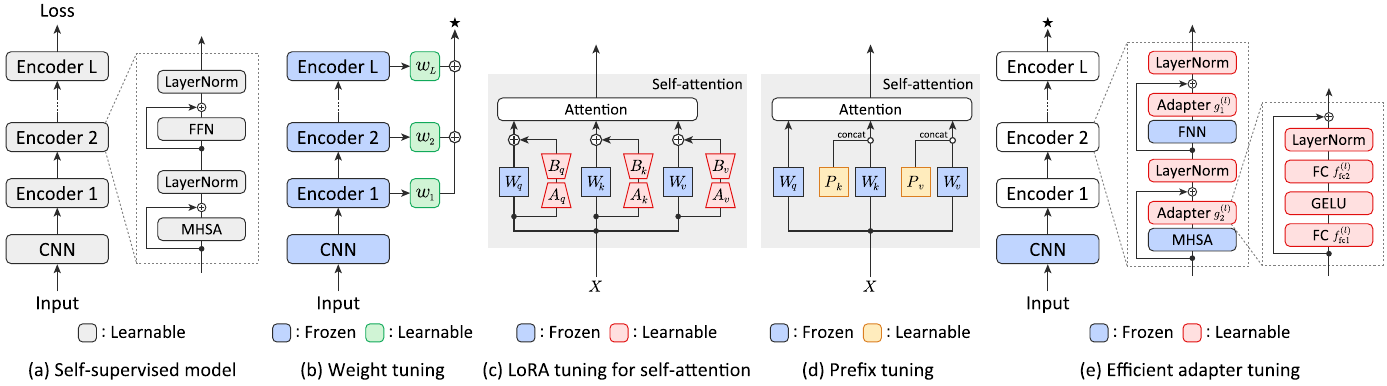}
\caption{Self-supervised model and conventional fine-tuning methods.
(a) Architecture of self-supervised model, which consists of a CNN encoder and $L$ transformer-based encoders. Vanilla transformer encoder, which consists of a multi-head self-attention (MHSA) module and a feedforward network (FFN) with LayerNorm and skip connections, is illustrated.
(b) Weight tuning applied to self-supervised model. It freezes all encoders and learns weights $w_{l}$ for each layer.
(c) LoRA tuning applied to self-attention module. It freezes weight matrices $W_{q}, W_{k}, W_{v}$ and injects learnable low-rank matrices $A_{q}, B_{q}, A_{k}, B_{k}, A_{v}, B_{v}$.
(d) Prefix tuning, which prepends learnable matrices $P_{k}$ and $P_{v}$ to the key and value matrices.
(e) Efficient adapter tuning applied to transformer-based encoder. It inserts two learnable adapters $g^{(l)}_{1}$ and $g^{(l)}_{2}$ to each layer, each of which involves two fully connected (FC) layers $f_{\text{fc1}}^{(l)}$ and $f_{\text{fc2}}^{(l)}$.
}
\label{fig:conventional}
\end{figure*}

%% file: fig/fig_overview.tex
\begin{figure*}
\centering
\includegraphics[width=\linewidth]{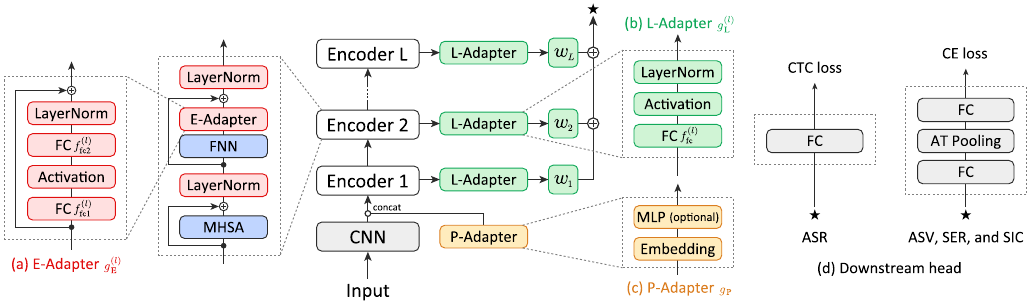}
\caption{Overview of ELP-Adapter tuning. Three types of adapters integrated into the self-supervised model.
(a) E-adapters (red) are inserted into each encoder layer to facilitate learning of fine-grained features for ASR.
(b) L-adapters (green) create paths from each encoder layer to the downstream head to extract non-linguistic features that are effective for ASV and SER.
(c) P-adapter (yellow) injects pseudo features into the output of the CNN encoder to further improve training effectiveness and efficiency.
(d) Minimal downstream heads (gray) are designed for each task to apply task-specific loss function.}
\label{fig_overview}
\end{figure*}

%% file: sec/3_proposed_method.tex
\section{Proposed adapter architecture}
\label{sec:method}
\input{fig/fig_adapter_modules}

This section presents ELP-adapter tuning, a novel fine-tuning method for various speech processing tasks.
Given a frozen self-supervised model (e.g., WavLM), ELP-adapter tuning incorporates three types of adapter, namely E-adapters, L-adapters, and a P-adapter, into the model, as shown in \cref{fig_overview}.
The E-adapters $g^{(l)}_{\hspace{1pt}\text{E}}$ are integrated into the transformer-based encoder layers. This helps to obtain fine-grained linguistic representations that are effective for speech recognition. The L-adapters $g^{(l)}_{\hspace{1pt}\text{L}}$ create paths from each encoder layer to the downstream head. This helps to extract features from intermediate encoder layers; such features are effective for emotion recognition and speaker verification. The P-adapter $g_{\hspace{1pt}\text{P}}$ appends learnable embeddings to input tokens to further enhance training effectiveness and efficiency.

The amount of storage required to store fine-tuned models is $O(N+K(M+H))$, where $K$ is the number of downstream tasks and $N$, $M$, and $H$ are the numbers of parameters of the frozen backbone model, learnable adapter modules, and downstream head, respectively.
Compared to full fine-tuning, for which the amount of storage required is $O(K (N+H))$, ELP-adapter is more efficient when $M \ll N$.
In practice, we need $M$ to be roughly 10 percent of $N$ to achieve performance comparable to that of full fine-tuning.
For example, with the WavLM backbone and our ELP-adapter modules, we have $N = 94.7$M and $M=9.52$M.
In the following, we provide detailed descriptions of each adapter module and downstream head.

\subsection{E-adapters}
The E-adapters $g^{(l)}_{\hspace{1pt}\text{E}}: \mathbb{R}^{n \times d} \to \mathbb{R}^{n \times d}$ are incorporated into each encoder layer to obtain fine-grained representations via fine-tuning as shown in \cref{fig_overview}(a).
Specifically, they are formulated as follows:
\begin{align}
\hat{Z}_{l} &= f_{\text{norm}}(f_{\text{mhsa}}^{(l)}(\hat{X}_{l-1}) + \hat{X}_{l-1}), \\
\label{eq:encea1}
\hat{X}_{l} &= f_{\text{norm}}(g^{(l)}_{\hspace{1pt}\text{E}}(f_{\text{ffn}}^{(l)}(\hat{Z}_{l})) + \hat{Z}_{l}), 
\end{align}
where $f_{\text{ffn}}^{(l)}$ is a frozen feedforward network, $f_{\text{mhsa}}^{(l)}$ is a frozen multi-head self-attention module, $f_{\text{norm}}$ is a normalization function, and $\hat{X}_{l}$ indicates the adapted output of the $l$-th encoder layer. Each E-adapter is given by
\begin{align}
g^{(l)}_{\hspace{1pt}\text{E}} (X) =
f_{\text{norm}}(f_{\text{fc2}}^{(l)}( \sigma( f_{\text{fc1}}^{(l)}(X)))) + X
\end{align}
where $f_{\text{fc1}}^{(l)}$ and $f_{\text{fc2}}^{(l)}$ are learnable fully connected layers and $\sigma$ is an activation function.
Compared to the conventional efficient adapter tuning in Eqs.~(\ref{eq:enca1}) and (\ref{eq:enca2}), the adapter module for MHSA is omitted. When the E-adapters are used with the L-adapters presented in the next subsection, this omission does not lead to a decrease in performance and improves parameter efficiency.
For activation function $\sigma$, the rectified linear unit (ReLU) is used for ASV and SER and GELU is used for ASR and IC.

\subsection{L-adapters}
The L-adapters make paths from each encoder layer to the downstream head to utilize the intermediate representations from the early phases of fine-tuning as shown in \cref{fig_overview}(b).
Let ${X}_{l}$ be the output of the $l$-th encoder layer.
The L-adapters $g^{(l)}_{\hspace{1pt}\text{L}}$ are applied to each ${X}_{l}$ to obtain adapted features as
\begin{align}
A_{l} = g^{(l)}_{\hspace{1pt}\text{L}}(X_{l})
\end{align}
for $l = 1, 2, \cdots, L$.
Each L-adapter is given by
\begin{align}
g^{(l)}_{\hspace{1pt}\text{L}}(X) = f_{\text{norm}}(\sigma( f_{\text{fc}}^{(l)}(X))),
\end{align}
where $f_{\text{fc}}^{(l)}$ is a learnable fully connected layer, $\sigma$ is an activation function, and $f_{\text{norm}}$ is a layer normalization function.
The weighted sum of the adapted features
\begin{align}
\label{qe:bara_ladapter}
\bar{A} = \sum_{l=1}^{L} w_{l} {A}_{l}
\end{align}
is then fed into the downstream head, where $w_{l} \in \mathbb{R}$ represents learnable weights.
This L-adapter is simpler than the conventional adapter module in Eq.~(\ref{nlpadapter}), resulting in better parameter efficiency.
The activation function $\sigma$ is the same as that used for the E-adapters.
The L-adapters are key components of our proposed method to cover various speech processing tasks, such as automatic speaker verification, where features extracted from lower layers are effective.

\subsection{P-adapter}
\label{sec:P-adapter}
Let $X_{0} \in \mathbb{R}^{n \times d}$ be the output of the CNN encoder, where $n$ is the length and $d$ is the dimension of each feature vector. The P-adapter injects pseudo features into it as shown in \cref{fig_overview}(c). We introduce four variants of P-adapters.

\subsubsection{Prefix P-adapter}

The prefix P-adapter $g_{\hspace{1pt}\text{pre}}$ prepends a new learnable matrix $P \in \mathbb{R}^{m \times d}$ as follows:
\begin{align}
g_{\hspace{1pt}\text{pre}}(X_{0}) = [P; X_{0}] \in \mathbb{R}^{(m+n) \times d},
\end{align}
where $[\hspace{3pt};\hspace{1pt}]$ indicates the concatenation operation.
Then, $Q_{0} = g_{\hspace{1pt}\text{pre}}(X_{0})$ is fed into the transformer encoders to obtain the encoder outputs $Q_{l}$ as
\begin{align}
Q_{l} = f_{\text{enc}}^{(l)}(Q_{l-1}),
\end{align}
where $f_{\text{enc}}^{(l)}$ indicates the $l$-th encoder. At the final layer, the first $m$ vectors corresponding to $P$ are omitted:
\begin{align}
\check{X}_{L} = g_{\hspace{1pt}\text{pre}}^{-1}(Q_{L}) \in \mathbb{R}^{n \times d}
\end{align}
where $g_{\hspace{1pt}\text{pre}}^{-1}$ is the inverse operation of $g_{\hspace{1pt}\text{pre}}$ for omitting the vectors.
This restores the sequence length to $n$.
The feature $\check{X}_{L}$ is fed into the downstream head.

When the P-adapter is utilized with the E-adapters, the E-adapters are applied to $Q_{l}$:
\begin{align}
\label{eq:encea1p}
\hat{Q}_{l} &= f_{\text{norm}}(g^{(l)}_{\hspace{1pt}\text{E}}(f_{\text{mlp}}^{(l)}(\hat{Z}_{l})) + \hat{Z}_{l}),\\
\hat{Z}_{l} &= f_{\text{norm}}(f_{\text{mhsa}}^{(l)}(\hat{Q}_{l-1}) + \hat{Q}_{l-1}).
\end{align}
When the P-adapter is utilized with the L-adapters, the inverse operation is inserted into each L-adapter as follows:
\begin{align}
A_{l} = g^{(l)}_{\hspace{1pt}\text{L}}(
g_{\hspace{1pt}\text{pre}}^{-1} (Q_{l})).
\end{align}

\subsubsection{Suffix P-adapter}
The suffix P-adapter $g_{\hspace{1pt}\text{pre}}$ appends a new learnable matrix $P \in \mathbb{R}^{m \times d}$ to $X_{0}$ as follows:
\begin{align}
g_{\hspace{1pt}\text{suf}}(X_{0}) = [X_{0}; P\hspace{1pt}] \in \mathbb{R}^{(n+m) \times d}.
\end{align}
This can be utilized with the E- and L-adapters in the same way as done for the prefix P-adapter.

\subsubsection{Nonlinear P-adapter}

To facilitate learning through pseudo features, the nonlinear P-adapter applies a small MLP $f_{\text{mlp}}$ to learnable embeddings $P$.
Specifically, we introduce two variants of the nonlinear P-adapter, namely prefix nonlinear P-adapter $g_{\hspace{1pt}\text{nl-pre}}$ and suffix nonlinear P-adapter $g_{\hspace{1pt}\text{nl-suf}}$, as follows:
\begin{align}
g_{\hspace{1pt}\text{nl-pre}}(X_{0}) &= [f_{\text{mlp}}(P); X_{0}] \in \mathbb{R}^{(m+n) \times d},\\
g_{\hspace{1pt}\text{nl-suf}}(X_{0}) &= [X_{0}; f_{\text{mlp}}(P)] \in \mathbb{R}^{(n+m) \times d}.
\end{align}

The best P-adapter configuration depends on the task, as we will discuss in \cref{sec:padapterdetails}.
We use the suffix P-adapter as the default P-adapter because it offers a good balance between performance and efficiency.

\subsection{Downstream head}
The downstream head is a minimal learnable module that is used to apply task-specific loss function.
This paper considers four tasks, ASR, ASV, SER and SIC, which belong to the four different aspects of speech~\cite{superb}: content, speaker, paralinguistics, and semantics, respectively.
As shown in \cref{fig_overview}(d), a single fully connected layer is used for ASR. A small network that consists of two fully connected layers with an average time pooling layer in between is used for ASV, SER, and SIC.

During fine-tuning, we also make all LayerNorm parameters learnable in the backbone self-supervised model, resulting in an addition of 0.037M learnable parameters. This approach is applied to all fine-tuning methods (weight tuning, prefix tuning, LoRA tuning, efficient adapter tuning, and our ELP-adapter tuning) in our experiments.

%% file: fig/fig_adapter_modules.tex
%% file: sec/4_asr.tex
\section{Automatic speech recognition}
\label{sec:ASR}

\input{tab/tab_asr_results}
\input{tab/tab_asr_ablation}

ASR aims to convert speech signals into text transcriptions.
For this task, speaker-independent features that distinguish phonemes often help improve performance.
In this section, we conduct experiments to demonstrate the effectiveness of ELP-adaptor tuning for the ASR task.

\subsection{Datasets and evaluation metrics}

The LibriLight dataset (train-10h) \cite{librilight} is used for training. It is a supervision training set that consists of 10 hours of audio data derived from open-source English audiobooks in the LibriVox project. The number of speakers is 24 (12 male, 12 female).
The LibriSpeech (dev-clean)~\cite{panayotov2015librispeech} dataset is used for testing. It consists of 5.4 hours of audio data with 40 speakers (20 male, 20 female).

The evaluation metric is the word error rate (WER), defined as
\begin{align}
\text{WER} = \frac{S+D+I}{N}
\end{align}
where $S$ is the number of substitutions, $D$ is the number of deletions, $I$ is the number of insertions, and $N$ is the number of words in the reference.

\subsection{Implementation details}
The downstream head for ASR consists of a single fully connected layer. CTC loss~\cite{graves2006ctc} is used as the loss function.
All models are fine-tuned with the Adam optimizer for $N_{\text{total}} = 34,600$ iterations with a batch size of 8.
The learning rate is scheduled with a linear warmup scheduler:
\begin{align}
\eta_t = 
\begin{cases} 
\eta_{0} + \frac{t}{N_{\text{warm}}} (\eta_{\text{max}} - \eta_0) & \text{if } t \leq N_{\text{warm}} \vspace{6pt}
\\
\eta_{\text{max}} - \left( \frac{t - N_{\text{warm}}}{N_{\text{total}} - N_{\text{warm}}} \right) \cdot (\eta_{\text{max}} - \eta_{0}) & \text{if } t > N_{\text{warm}}
\end{cases}
\end{align}
where $t$ is the time step,
$N_{\text{warm}} = 5,000$ is the number of warmup steps, $\eta_{0} = 10^{-7}$ is the initial and final learning rate, and $\eta_{\text{max}}$ is the maximum learning rate after warmup.
For each fine-tuning method, the best learning rate for $\eta_{\text{max}}$ is chosen from $\{1.0 \times 10^{-3}, 5.0 \times 10^{-4}, 1.0 \times 10^{-4}, 5.0 \times 10^{-5}, 1.0 \times 10^{-5}\}$.

\subsection{Comparison with conventional methods}

Table~\ref{tab:asr_results} compares ELP-adapter tuning with the conventional fine-tuning methods described in Section~\ref{sec:fine-tuning_methods}.
As shown, our method outperformed the conventional methods for the five self-supervised models while having fewer learnable parameters than that for the conventional efficient adapter tuning.
This shows the effectiveness and parameter efficiency of ELP-adapter tuning.
It is also worth noting that ELP-adapter tuning achieved WERs lower than those obtained by full fine-tuning for two models (HuBERT and WavLM+).
This is because ELP-adapter allows for quick adaptation while avoiding overfitting.

Regarding the self-supervised models, WavLM showed the best performance among the four models pre-trained on LibriSpeech (wav2vec2.0, HuBERT, ContentVec, and WavLM), with the exception of the prefix tuning result. This is because the gated relative position bias used in WavLM is particularly effective for ASR.
With prefix tuning, wav2vec2.0 provided the best fit. This is because when adding new elements to the key and value matrices of attention, a simpler architecture works better.
In addition, WavLM+ outperformed WavLM in all cases. This shows that increasing the amount of pre-training data improves performance, regardless of the fine-tuning method.
\input{fig/fig_tradeoff_asr}

\subsection{Ablation study}
Table~\ref{tab:asr_ablation} shows the results of the ablation study for various adapter types.
As shown, E-adapter tuning outperformed L-adapter tuning for the five self-supervised models.
This indicates that the adaptation of encoders plays a more crucial role in ASR than the fusion of outputs from multiple layers via L-adapters.
For ASR, features from layers closer to the last layer, which are often speaker-independent phoneme-level features, contribute to the performance improvement. Therefore, the insertion of E-adapters into a series of encoder layers to adapt these features was effective.

The combination of E-adapters and L-adapters reduced the WER for all models. The addition of P-adapter further reduced the average performance with a small increase in the number of learnable parameters. This demonstrates the effectiveness of the proposed combination of three types of adapter.

\subsection{Trade-off analysis}
We performed experiments in which the numbers of layers used to fine-tune and insert adapters was varied to find cases where a
smaller number of parameters would perform well.
Figure~\ref{fig:tradeoff_asr} compares the results obtained with full fine-tuning, conventional efficient adapter tuning, and the proposed ELP-adapter tuning.
As shown, all error curves decrease quickly, showing that adjustments of only the top three layers are sufficient.
This suggests that encoders in the lower layers are already effective at extracting features for ASR and that freezing them to avoid overfitting can enhance performance.
We also confirmed that our method had the best performance in all cases.


%% file: tab/tab_asr_results.tex
\def\confidence{Confidence intervals were obtained by repeating each experiment five times with different seeds.}
\def\confidenceshort{Confidence intervals were obtained by repeating each experiment five times with different seeds.}
\tabresulttemplate
{
\setlength{\tabcolsep}{8.7pt}
\caption{Comparison of fine-tuning methods on ASR task. Word error rates (\%) on LibriSpeech (dev set) are reported. Best and second best results are highlighted in bold and underlined, respectively.
\confidence}
\label{tab:asr_results}}
{ $24.92_{\pm 0.169}$ & $28.89_{\pm 0.081}$ & $33.98_{\pm 0.088}$& $24.17_{\pm 0.151}$ &$21.59_{\pm 0.387}$ & $26.71_{\pm 0.092}$}
{ $18.38_{\pm 0.267}$ & $22.64_{\pm 0.494}$ & $27.23_{\pm 0.165}$& $28.21_{\pm 0.533}$ &$25.81_{\pm 0.531}$ & $24.45_{\pm 0.190}$}
{ $11.51_{\pm 0.099}$ & $12.30_{\pm 0.140}$ & $16.56_{\pm 0.221}$& $10.64_{\pm 0.089}$ &$9.85_{\pm 0.144}$ & $12.17_{\pm 0.065}$}
{ $\underline{9.91}_{\pm 0.073}$ & $\underline{9.94}_{\pm 0.084}$ & $\underline{12.97}_{\pm 0.112}$& $\underline{9.27}_{\pm 0.098}$ &$\underline{8.59}_{\pm 0.059}$ & $\underline{10.13}_{\pm 0.038}$}
{$\bm{9.30}_{\pm 0.034}$ & $\bm{9.20}_{\pm 0.101}$ & $\bm{12.07}_{\pm 0.059}$ & $\bm{8.53}_{\pm 0.012}$ & $\bm{7.85}_{\pm 0.072}$ & $\bm{9.39}_{\pm 0.028}$}
{ $9.26_{\pm 0.132}$ & $9.30_{\pm 0.085}$ & $12.00_{\pm 0.044}$& $8.50_{\pm 0.085}$ &$8.02_{\pm 0.070}$ & $9.41_{\pm 0.0393}$}


%% file: tab/tab_asr_ablation.tex
\tabalbationtemplate
{\caption{
Ablation study on ASR task.
Word error rates (\%) on LibriSpeech (dev set) are reported. Best results are highlighted in bold. \confidenceshort}
\label{tab:asr_ablation}}
{$\bm{9.30}_{\pm 0.034}$ & $9.20_{\pm 0.101}$ & $\bm{12.07}_{\pm 0.059}$ & $8.53_{\pm 0.012}$ & $\bm{7.85}_{\pm 0.072}$ & $\bm{9.39}_{\pm 0.028}$}
{$27.48_{\pm 0.639}$ & $32.53_{\pm 0.201}$ & $35.49_{\pm 0.711}$ & $26.22_{\pm 0.196}$ & $23.68_{\pm 0.297}$ & $29.08_{\pm 0.208}$}
{ $10.06_{\pm 0.135}$ & $10.41_{\pm 0.136}$ & $13.91_{\pm 0.100}$& $9.62_{\pm 0.111}$ &$9.09_{\pm 0.052}$ & $10.62_{\pm 0.050}$}
{ $12.21_{\pm 0.135}$ & $11.59_{\pm 0.028}$ & $14.78_{\pm 0.074}$& $10.45_{\pm 0.093}$ &$9.50_{\pm 0.089}$ & $11.72_{\pm 0.040}$}
{ $9.59_{\pm 0.495}$ & $\bm{9.16}_{\pm 0.104}$ & $12.12_{\pm 0.106}$& $\bm{8.43}_{\pm 0.098}$ &$7.86_{\pm 0.061}$ & $9.43_{\pm 0.106}$}

%% file: fig/fig_tradeoff_asr.tex
\def\wavlmisused{The WavLM model was used as a backbone model.}
\def\wavlmisuseds{The WavLM model is used as a backbone model.}
\begin{figure}
\centering
\includegraphics[width=\linewidth]{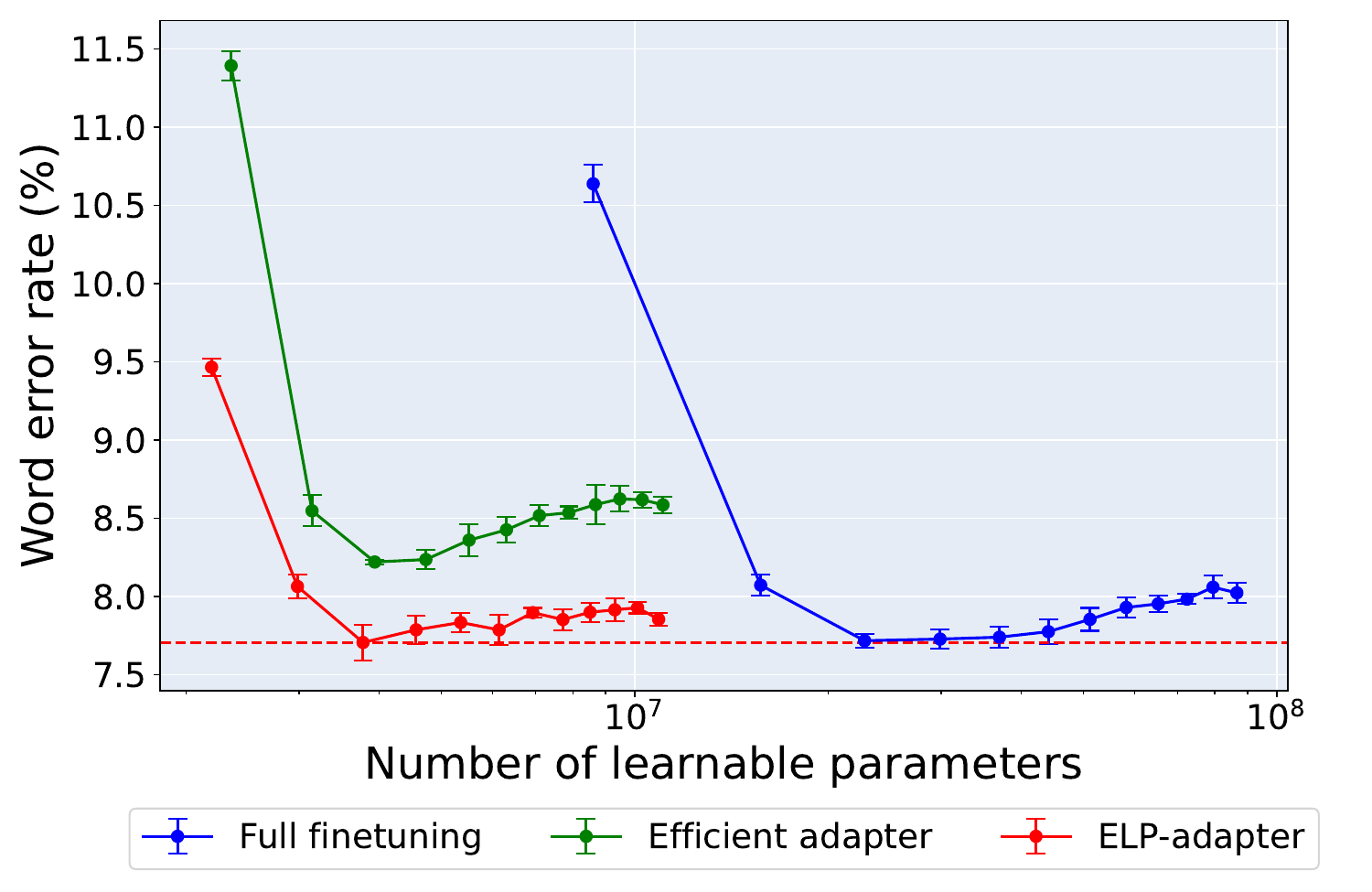}
\caption{Trade-off between number of learnable parameters and ASR performance in terms of WER with number of frozen layers varying from 1 to 12. \wavlmisused}
\label{fig:tradeoff_asr}
\end{figure}

%% file: sec/5_asv.tex
\section{Automatic speaker verification}
\label{sec:ASV}

ASV aims to verify the identity of speakers.
Given an enrollment utterance and a test utterance, the goal is to determine whether these two utterances are from the same speaker. 
This paper focuses on text-independent speaker verification, which has no constraints on speech content.
For this task, speaker-dependent features that are robust to changes in speech content and background noise often help improve performance.
This section applies ELP-adapter tuning to the ASV task.

\input{tab/tab_asv_results}
\input{tab/tab_asv_ablation}

\subsection{Datasets and evaluation metrics}
The VoxCeleb1 dataset \cite{voxceleb} is used for training and testing.
It consists of 351 hours of audio data extracted from videos uploaded to YouTube.
The training set consists of 148,642 audio utterances from 1,211 speakers.
The test set consists of 37,611 trials built from 4,874 audio utterances from 40 speakers.

The evaluation metric is the equal error rate (EER), which is the error rate at which the false alarm rate (FAR) and the false rejection rate (FRR) are equal. Here, FAR and FRR are given by
\begin{align}
\mathrm{FAR} = \frac{\mathrm{FP}}{\mathrm{FP}+\mathrm{TP}},~
\mathrm{FRR} = \frac{\mathrm{FN}}{\mathrm{TP}+\mathrm{FN}},
\end{align}
where $\mathrm{TP}$ is the number of true positives, 
$\mathrm{FP}$ is the number of false positives, and 
$\mathrm{FN}$ is the number of false negatives.

\subsection{Implementation details}
The downstream head for ASV consists of a small MLP, which has two fully connected layers and an average time pooling layer in between.
The number of hidden units is set to 768.
The cross-entropy loss with speaker ID labels is used as the loss function, by which models learn to classify speakers.
All models are fine-tuned with the Adam optimizer for 20.8k iterations.
The batch size is determined adaptively at each iteration to fit as much data as possible into 16~GB of GPU memory.
The learning rate is scheduled with the linear warmup scheduler with $N_{\text{warm}} = 5,000$.
In the verification phase, speaker embeddings are extracted from the average time pooling layer by omitting the final fully connected layer.
For each trial, the cosine similarity between the speaker embeddings for the enrollment and test utterances is measured to determine whether the two utterances are from the same speaker, with the threshold set such that FAR and FRR are equal to compute EER.
The adaptive s-norm~\cite{snorm1, snorm2} is applied to normalize these trial scores.

\subsection{Comparison with conventional methods}

\input{fig/fig_tradeoff_asv}

Table~\ref{tab:asv_results} compares ELP-adapter tuning with the four conventional fine-tuning methods.
As shown, our method has the best performance for HuBERT, ContentVec, WavLM, and WavLM+.
This advantage is derived from the use of L-adapters, which connect the outputs of each layer to the downstream head. As discussed in the ASR experiments, features in the upper layers tend to be speaker-independent. Therefore, connecting the lower layers to the downstream head helps to improve the extraction of speaker-dependent features, which is essential for ASV.
With wav2vec 2.0, speaker information and prosodic information could be entangled even in the upper layers, making simple encoder adaptation with conventional efficient adapter tuning the best solution.

With full fine-tuning, HuBERT performed the best and WavLM+ performed the worst, in contrast to the results for ASR. This indicates that the features of WavLM+, especially those of the last layer, are highly speaker-independent.
ELP-adapter tuning effectively addresses this limitation, achieving the best performance with WavLM+.

\subsection{Ablation study}
Table~\ref{tab:asv_ablation} shows the results of the ablation study.
As shown, L-adapter tuning outperformed E-adapter tuning for HuBERT, ContentVec, WavLM, and WavLM+.
Notably, L-adapter tuning significantly improves the performance of WavLM+, with a 1.96 point decrease in EER.
This supports the above discussion about the effectiveness of L-adapters for ASV.
The combination of E-adapters and L-adapters improved performance for all models, and the addition of P-adapter further improved performance for four of the five models (Wav2vec2.0, ContentVec, WavLM and WavLM+).
This result is consistent with that for the ASR task.

\subsection{Trade-off analysis}

Figure~\ref{fig:tradeoff_asv} shows the results obtained with various numbers of fine-tuned layers for full fine-tuning, conventional efficient adapter tuning, and proposed ELP-adapter tuning.
As shown, EER decreases as the number of fine-tuned layers increases.
In contrast to the ASR results in Figure~\ref{fig:tradeoff_asr},
fine-tuning lower layers improves performance because these layers facilitate the extraction of speaker-dependent non-linguistic features.
We also confirmed that our method had the best performance in all cases.

%% file: tab/tab_asv_results.tex
\tabresulttemplate
{
\setlength{\tabcolsep}{10pt}
\caption{Comparison of fine-tuning methods on ASV task. Equal error rates (\%) on VoxCeleb1 are reported. Best and second best results are highlighted in bold and underlined, respectively. \confidenceshort}\label{tab:asv_results}}
{ $5.76_{\pm 0.124}$ & $5.71_{\pm 0.077}$ & $6.27_{\pm 0.095}$& $6.44_{\pm 0.254}$ &$5.23_{\pm 0.062}$ & $5.88_{\pm 0.062}$}
{$4.96_{\pm 0.180}$ & $4.91_{\pm 0.151}$ & $6.73_{\pm 0.198}$& $6.36_{\pm 0.198}$ & $8.07_{\pm 0.571}$ & $6.21_{\pm 0.136}$}
{ $4.38_{\pm 0.055}$ & $4.06_{\pm 0.137}$ & $4.93_{\pm 0.104}$& $4.63_{\pm 0.119}$ &$\underline{3.98}_{\pm 0.283}$ & $4.40_{\pm 0.071}$}
{ $\bm{3.38}_{\pm 0.145}$ & $\underline{3.17}_{\pm 0.131}$ & $\underline{3.82}_{\pm 0.099}$& $\underline{3.59}_{\pm 0.048}$ &$\underline{3.98}_{\pm 0.361}$ & $\underline{3.69}_{\pm 0.085}$}
{$\underline{3.53}_{\pm 0.037}$ & 
$\bm{3.16}_{\pm 0.022}$ &
$\bm{3.42}_{\pm 0.092}$ &
$\bm{3.21}_{\pm 0.069}$ &
$\bm{2.57}_{\pm 0.129}$ & $\bm{3.18}_{\pm 0.035}$ }
{ $3.57_{\pm 0.335}$ & $3.06_{\pm 0.094}$ & $3.84_{\pm 0.140}$& $3.66_{\pm 0.265}$ &$4.27_{\pm 0.194}$ & $3.68_{\pm 0.100}$~}

%% file: tab/tab_asv_ablation.tex
\tabalbationtemplate{
\setlength{\tabcolsep}{11.35pt}
\caption{Ablation study on ASV task.
Equal error rates (\%) on VoxCeleb1 are reported. Best results are highlighted in bold. \confidenceshort}
\label{tab:asv_ablation}
}
{$\bm{3.53}_{\pm 0.037}$ & 
$3.17_{\pm 0.022}$ &
$\bm{3.42}_{\pm 0.092}$ &
$\bm{3.21}_{\pm 0.069}$ &
$\bm{2.57}_{\pm 0.129}$ & $\bm{3.18}_{\pm 0.035}$}
{$5.27_{\pm 0.057}$ & 	$5.37_{\pm 0.130}$ & 	$6.96_{\pm 0.160}$ & 	$6.05_{\pm 0.020}$ & 	$5.62_{\pm 0.114}$ & $5.85_{\pm 0.049}$}
{ $3.34_{\pm 0.070}$ & $3.66_{\pm 0.135}$ & $4.56_{\pm 0.224}$& $3.71_{\pm 0.059}$ &$4.65_{\pm 0.302}$ & $3.98_{\pm 0.082}$}
{ $3.90_{\pm 0.038}$ & $3.33_{\pm 0.038}$ & $4.00_{\pm 0.036}$& $3.50_{\pm 0.078}$ &$2.74_{\pm 0.097}$ & $3.48_{\pm 0.031}$}
{ $3.59_{\pm 0.187}$ & $\bm{3.06}_{\pm 0.075}$ & $3.47_{\pm 0.069}$& $3.33_{\pm 0.083}$ &$2.62_{\pm 0.045}$ & $3.21_{\pm 0.047}$}

%% file: fig/fig_tradeoff_asv.tex
\begin{figure}
\centering
\includegraphics[width=\linewidth]{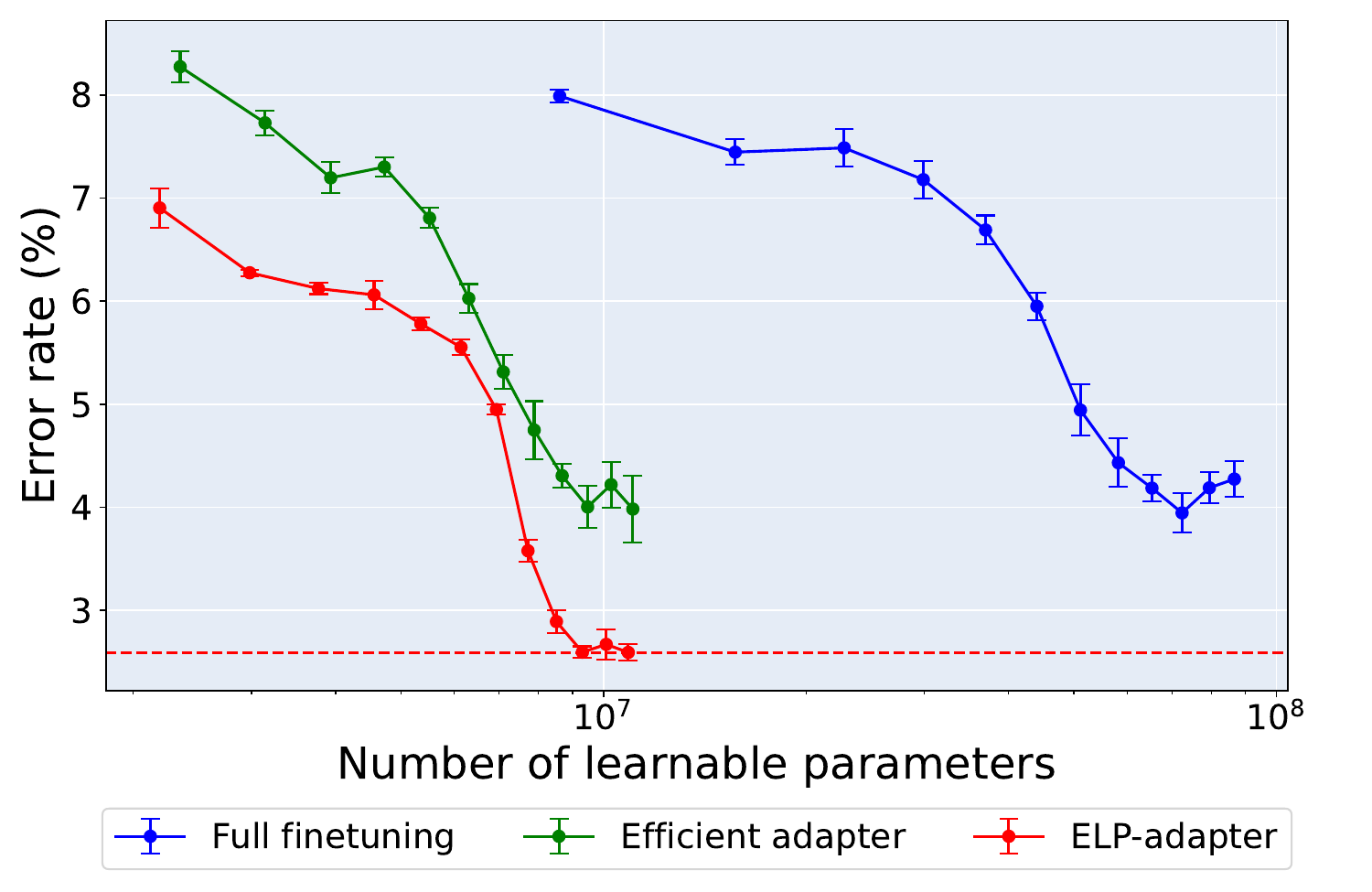}
\caption{Trade-off between number of learnable parameters and ASV performance in terms of EER. \wavlmisuseds}
\label{fig:tradeoff_asv}
\end{figure}

%% file: sec/6_ser.tex
\input{tab/tab_ser_results}
\input{tab/tab_ser_ablation}

\section{Speech emotion recognition}
\label{sec:SER}

SER aims to identify the affect of the speaker based on audio utterances. It is often formulated as a classification problem, where the goal is to classify the input audio utterance into predefined emotion classes.
For this task, audio features that effectively capture emotional cues in speech, such as tone, pitch, and rhythm, are crucial for enhancing accuracy.
This section applies ELP-adapter tuning to the SER task.

\subsection{Datasets and evaluation metrics}
The IEMOCAP dataset~\cite{iemocap} is used for training and testing.
It consists of 12 hours of audio data collected from 10 actors (5 male, 5 female) performing scripted and spontaneous dialogues. 
Following previous studies~\cite{erpaper}, four emotion categories, namely ``neutral'', ``happy'', ``sad'', and ``angry'', are used for evaluation, where ``excited'' is merged into ``happy''.

The evaluation metric is the error rate (ER), which is given by
\begin{align}
\mathrm{ER} = 1 - \frac{1}{C}\sum_{i=1}^{C} \mathrm{ACC}_{i},
\end{align}
where $C = 4$ is the number of emotion classes and $\mathrm{ACC}_{i}$ is the accuracy for the $i$-th class.
Five-fold cross-validation is performed to measure ER.

\subsection{Implementation details}
The downstream head for SER consists of a small MLP, which has two fully connected layers and an average time pooling layer in between.
The number of hidden units is set to 256.
The cross-entropy loss is used as a loss function.
All models are fine-tuned with the Adam optimizer for 2,750 iterations.
The batch size is set to 32.
The learning rate is scheduled with a step scheduler, which is given by
\begin{align}
\eta_{t} = \eta_{0} \cdot (\gamma ^ {\lfloor t / s \rfloor})
\end{align}
where $\gamma = 0.1$ and $s = 10$.

\subsection{Comparison with conventional methods}

Table~\ref{tab:ser_results}
shows that ELP-adapter tuning outperforms the four conventional fine-tuning methods.
For SER, it is necessary to comprehensively extract prosodic information such as tone, pitch, and rhythm.
Similar to the case for ASV, L-adapters helped to extract non-linguistic features from lower encoder layers.

Because most self-supervised models are trained to find hidden audio units in an unsupervised manner on clean non-emotional speech data, which often leads to the formation of units capable of distinguishing phonemes but not emotions,
features extracted from frozen self-supervised models are not always effective for SER.
Nevertheless, ELP-adapter with WavLM+ achieved the best performance among all methods.
This highlights the potential of adapter-based fine-tuning to handle complex tasks such as SER.

\subsection{Ablation study}
Table~\ref{tab:ser_ablation} shows the results of the ablation study.
As shown, L-adapter tuning outperformed E-adapter tuning for all models.
Similar to ASV, this result indicates that features extracted from lower layers are useful for SER because they often represent low-level features such as pitch and tone, which help to discriminate emotions.
The combination of multiple types of adapter further improved the performance.
This result is consistent with those for ASR and ASV.

\subsection{Trade-off analysis}
Figure~\ref{fig:tradeoff_ser} shows the results obtained with various numbers of fine-tuned layers.
ER decreases as the number of fine-tuned layers increases.
This tendency is similar to that observed for ASV, but unlike for ASV, the error curve does not exhibit a sharp bend. This indicates that the high-level linguistic features in upper layers effective for ASR are also beneficial for SER. It was also confirmed that the proposed method outperforms the conventional methods in all cases.
\input{fig/fig_tradeoff_ser}

%% file: tab/tab_ser_results.tex
\tabresulttemplate
{
\setlength{\tabcolsep}{7.75pt}\caption{Comparison of fine-tuning methods on SER task. Equal rates (\%) on IEMOCAP are reported. Best and second best results are highlighted in bold and underlined, respectively. \confidenceshort}\label{tab:ser_results}}
{ $27.76_{\pm 0.370}$ & $27.48_{\pm 0.288}$ & $29.77_{\pm 0.495}$& $29.40_{\pm 0.637}$ &$28.47_{\pm 0.571}$ & $28.58_{\pm 0.219}$}
{ $28.58_{\pm 0.683}$ & $28.85_{\pm 0.431}$ & $29.32_{\pm 0.359}$& $35.77_{\pm 0.480}$ &$33.12_{\pm 0.531}$ & $31.13_{\pm 0.227}$}
{ $\underline{25.48}_{\pm 0.234}$ & $25.34_{\pm 0.612}$ & $27.13_{\pm 0.572}$& $25.35_{\pm 0.342}$ &$24.22_{\pm 0.570}$ & $25.50_{\pm 0.219}$}
{ $25.63_{\pm 0.256}$ & $\underline{22.59}_{\pm 0.486}$ & $\underline{24.34}_{\pm 0.369}$& $\underline{23.58}_{\pm 4.689}$ &$\underline{21.56}_{\pm 0.240}$ & $\underline{23.54}_{\pm 0.197}$}
{$\bm{21.74}_{\pm 0.131}$ & $\bm{22.34}_{\pm 0.233}$ & $\bm{22.93}_{\pm 0.765}$ & $\bm{20.45}_{\pm 0.406}$ & 
$\bm{19.88}_{\pm 0.368}$ & $\bm{21.47}_{\pm 0.196}$}
{ $22.70_{\pm 0.319}$ & $20.73_{\pm 0.379}$ & $24.11_{\pm 0.549}$ & $20.25_{\pm 0.400}$ &$20.39_{\pm 0.125}$ & $21.64_{\pm 0.1700}$}

%% file: tab/tab_ser_ablation.tex
\tabalbationtemplate{
\setlength{\tabcolsep}{9.1pt}
\caption{Ablation study on SER task.
Equal error rates (\%) on IEMOCAP are reported. Best results are highlighted in bold. \confidenceshort}
\label{tab:ser_ablation}
}
{$\bm{21.74}_{\pm 0.131}$ & $22.34_{\pm 0.233}$ & $22.93_{\pm 0.765}$ & $\bm{20.45}_{\pm 0.406}$ & 
$\bm{19.88}_{\pm 0.368}$ & $\bm{21.47}_{\pm 0.196}$}
{$30.17_{\pm 0.408}$ & $31.54_{\pm 0.433}$ & 	$32.29_{\pm 0.496}$ & $32.14_{\pm 0.935}$ & 	$30.14_{\pm 1.106}$ & $31.26_{\pm 0.329}$}
{ $27.13_{\pm 0.969}$ & $28.38_{\pm 0.867}$ & $31.11_{\pm 0.807}$& $29.94_{\pm 1.445}$ &$29.41_{\pm 0.882}$ & $29.19_{\pm 0.456}$}
{ $25.04_{\pm 0.834}$ & $24.94_{\pm 1.058}$ & $24.99_{\pm 0.680}$& $24.49_{\pm 0.902}$ &$21.12_{\pm 0.522}$ & $24.50_{\pm 0.392}$}
{ $22.11_{\pm 0.758}$ & $\bm{22.01}_{\pm 0.352}$ & $\bm{22.87}_{\pm 0.312}$& $21.27_{\pm 0.331}$ &$20.26_{\pm 0.652}$ & $21.70_{\pm 0.231}$}

%% file: fig/fig_tradeoff_ser.tex
\begin{figure}
\centering
\includegraphics[width=\linewidth]{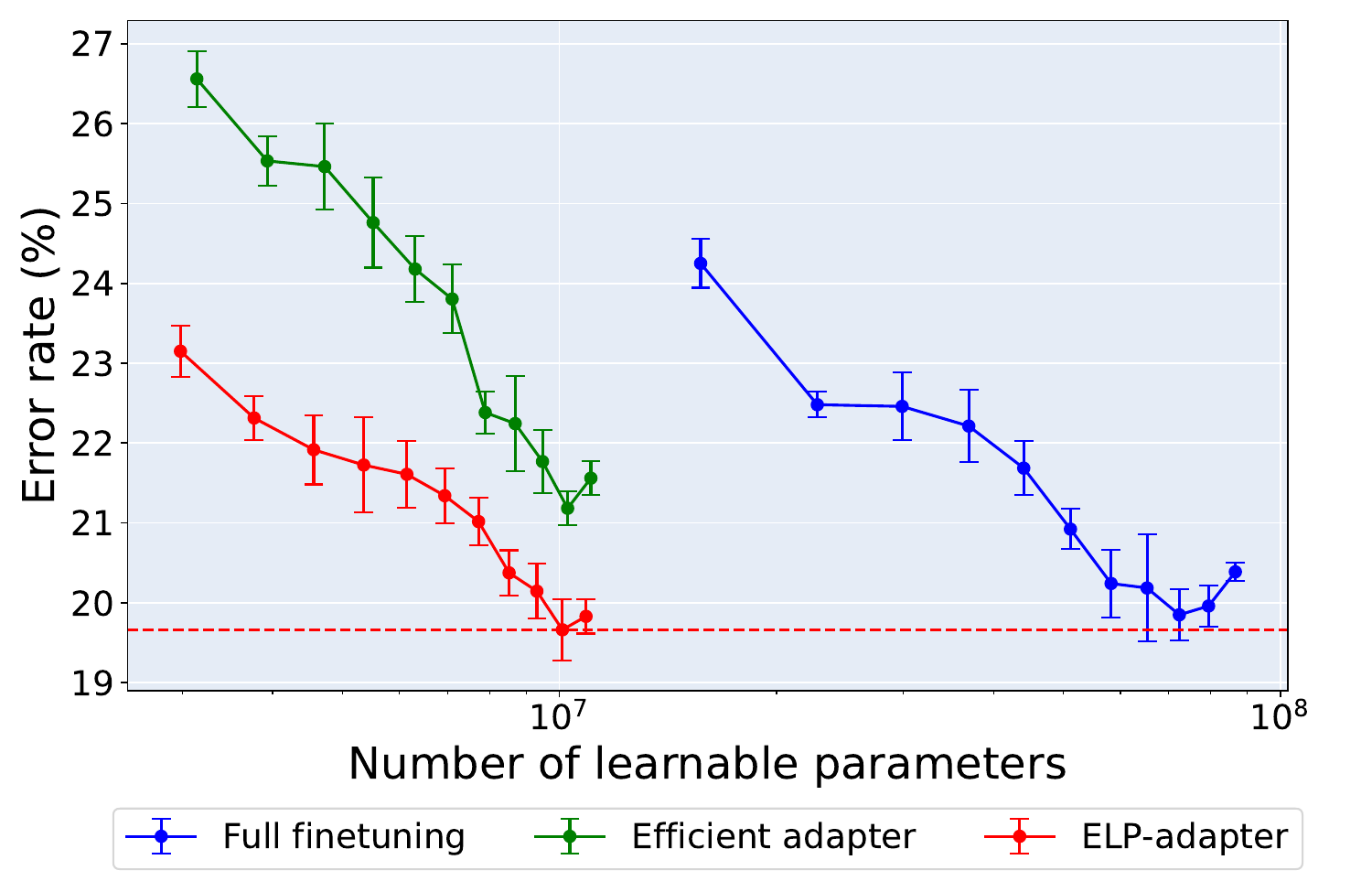}
\caption{Trade-off between number of learnable parameters and SER performance in terms of ER. \wavlmisuseds}
\label{fig:tradeoff_ser}
\end{figure}

%% file: sec/7_ic.tex
\section{Speech intent classification}
\label{sec:SIC}
\input{tab/tab_ic_results}
\input{tab/tab_ic_ablation}

SIC aims to identify the purpose behind an audio input. It requires understanding and categorizing the intent into predefined classes.
For this task, features that capture semantic information often play a critical role in improving performance. In this section, we apply ELP-adaptor tuning to the IC task.

\subsection{Datasets and evaluation metrics}
The Fluent Speech Commands dataset~\cite{speechcommand} is used for training and testing.
It consists of simple voice assistant commands with 30,043 English audio utterances from 97 speakers.
Each utterance is labeled with three slots: ``action'', ``object'', and ``location''.
A set of classes is predefined for each slot.
Specifically, there are 6 action classes, 14 object classes, and 4 location classes:
{
\setlength{\topsep}{-1pt}
\setlength{\FrameSep}{3pt}
\setlength{\parindent}{0pt}
\begin{framed}
\small
\noindent
Action: 1) activate, 2) bring, 3) change language, 4) deactivate, 5)~decrease, 6) increase
\end{framed}
\begin{framed}
\small
\noindent 
Object: 1) Chinese, 2) English, 3) German, 4) Korean, 5) heat, 6)~juice, 7) lamp, 8) lights, 9) music, 10) newspaper, 11) none, 12)~shoes, 13) socks, 14) volume
\end{framed}
\begin{framed}
\small
\noindent 
Location: 1) bedroom, 2) kitchen, 3) washroom, 4) none
\end{framed}
}

The evaluation measure is ER, defined as $1.0 - \mathrm{ACC}$, where $\mathrm{ACC}$ is the accuracy computed with true positives defined as the correct classifications with respect to all three slots.

\subsection{Comparison with conventional methods}

Table~\ref{tab:ic_results} compares ELP-adapter tuning with the four conventional fine-tuning methods on the SIC task.
Most methods achieved an ER of less than 1.0\%. Conventional efficient adapter tuning, full fine-tuning, and ELP-adapter tuning had comparable performance.
The absence of significant differences between these three methods indicates that the SIC task is relatively simple compared to tasks such as ASR and ASV, suggesting that tuning with only encoder adapters may be sufficient.

\subsection{Ablation study}
Table~\ref{tab:ic_ablation} shows the results of the ablation study.
In contrast to ASR, ASV, and SER, the combination of the three types of adapter was the most effective only for two models (HuBERT and ContentVec).
This is because minimizing loss on SIC is relatively easier than the other tasks, and combining three types of adapters is redundant.
This paper aimed to propose an adapter tuning method that is effective for various speech processing tasks.
However, automatic pruning of unnecessary adapters is also an interesting topic for future research.

\subsection{Trade-off analysis}

Figure~\ref{fig:tradeoff_sic} shows the results obtained with various numbers of fine-tuned layers.
Tendencies similar to those for ASR can be seen, where fine-tuning the upper layers well reduces ER and fine-tuning all layers results in overfitting.
This is because linguistic information and high-level semantic information extracted from the upper layers are crucial for understanding intent behind audio inputs.
The conventional efficient adapter tuning had the best performance with a small number of fine-tuned layers (the best performance is archived with five layers).
This confirms that tuning only encoder adapters is the most efficient method for this task.
\input{fig/fig_tradeoff_sic}

%% file: tab/tab_ic_results.tex
\tabresulttemplate
{
\setlength{\tabcolsep}{9.55pt}\caption{Comparison of fine-tuning methods on IC task. Equal rates (\%) on Fluent Speech Commands are reported. Best and second best results are highlighted in bold and underlined, respectively. \confidenceshort}\label{tab:ic_results}}
{ $1.79_{\pm 0.800}$ & $0.70_{\pm 0.094}$ & $1.03_{\pm 0.202}$& $1.68_{\pm 0.276}$ &$0.45_{\pm 0.101}$ & $1.13_{\pm 0.176}$}
{ $0.68_{\pm 0.144}$ & $0.51_{\pm 0.057}$ & $0.53_{\pm 0.062}$& $1.84_{\pm 0.232}$ &$3.17_{\pm 0.486}$ & $1.35_{\pm 0.113}$}
{ $0.71_{\pm 0.212}$ & $0.53_{\pm 0.067}$ & $0.47_{\pm 0.068}$& $0.42_{\pm 0.034}$ &$0.41_{\pm 0.047}$ & $0.51_{\pm 0.048}$}
{ $\bm{0.41}_{\pm 0.051}$ & $\underline{0.43}_{\pm 0.088}$ & $\underline{0.39}_{\pm 0.063}$& $\bm{0.36}_{\pm 0.078}$ &$\bm{0.35}_{\pm 0.040}$ & $\underline{0.39}_{\pm 0.030}$}
{$\underline{0.44}_{\pm 0.047}$&$\bm{0.34}_{\pm 0.067}$
&$\bm{0.32}_{\pm 0.067}$&$\underline{0.40}_{\pm 0.068}$&$\underline{0.39}_{\pm 0.050}$ & $\bm{0.38}_{\pm 0.027}$}
{ $0.41_{\pm 0.043}$ & $0.37_{\pm 0.019}$ & $0.37_{\pm 0.032}$& $0.36_{\pm 0.080}$ &$0.41_{\pm 0.076}$ & $0.38_{\pm 0.025}$}

%% file: tab/tab_ic_ablation.tex
\tabalbationtemplate{
\setlength{\tabcolsep}{10.9pt}
\caption{Ablation study on IC task. Equal rates (\%) on Fluent Speech Commands are reported. Best results are highlighted in bold. \confidenceshort}
\label{tab:ic_ablation}
}
{$0.44_{\pm 0.047}$&$\bm{0.34}_{\pm 0.067}$
&$\bm{0.32}_{\pm 0.067}$&$0.40_{\pm 0.068}$&$0.39_{\pm 0.050}$ & $\bm{0.38}_{\pm 0.027}$}
{
$0.90_{\pm 0.225}$ & 	$0.65_{\pm 0.135}$ & 	$0.52_{\pm 0.066}$ & 	$0.58_{\pm 0.046}$ & 	$0.80_{\pm 0.300}$ & $0.69_{\pm 0.081}$
}
{ $0.53_{\pm 0.179}$ & $\bm{0.34}_{\pm 0.046}$ & $0.37_{\pm 0.067}$& $\bm{0.37}_{\pm 0.063}$ &$0.35_{\pm 0.022}$ & $0.39_{\pm 0.041}$}
{ $0.73_{\pm 0.215}$ & $0.48_{\pm 0.029}$ & $0.42_{\pm 0.029}$& $0.40_{\pm 0.043}$ &$\bm{0.33}_{\pm 0.014}$ & $0.47_{\pm 0.045}$}
{ $\bm{0.41}_{\pm 0.051}$ & $0.36_{\pm 0.090}$ & $0.38_{\pm 0.069}$& $0.39_{\pm 0.055}$ &$0.36_{\pm 0.043}$ & $\bm{0.38}_{\pm 0.029}$}

%% file: fig/fig_tradeoff_sic.tex
\begin{figure}
\centering
\includegraphics[width=\linewidth]{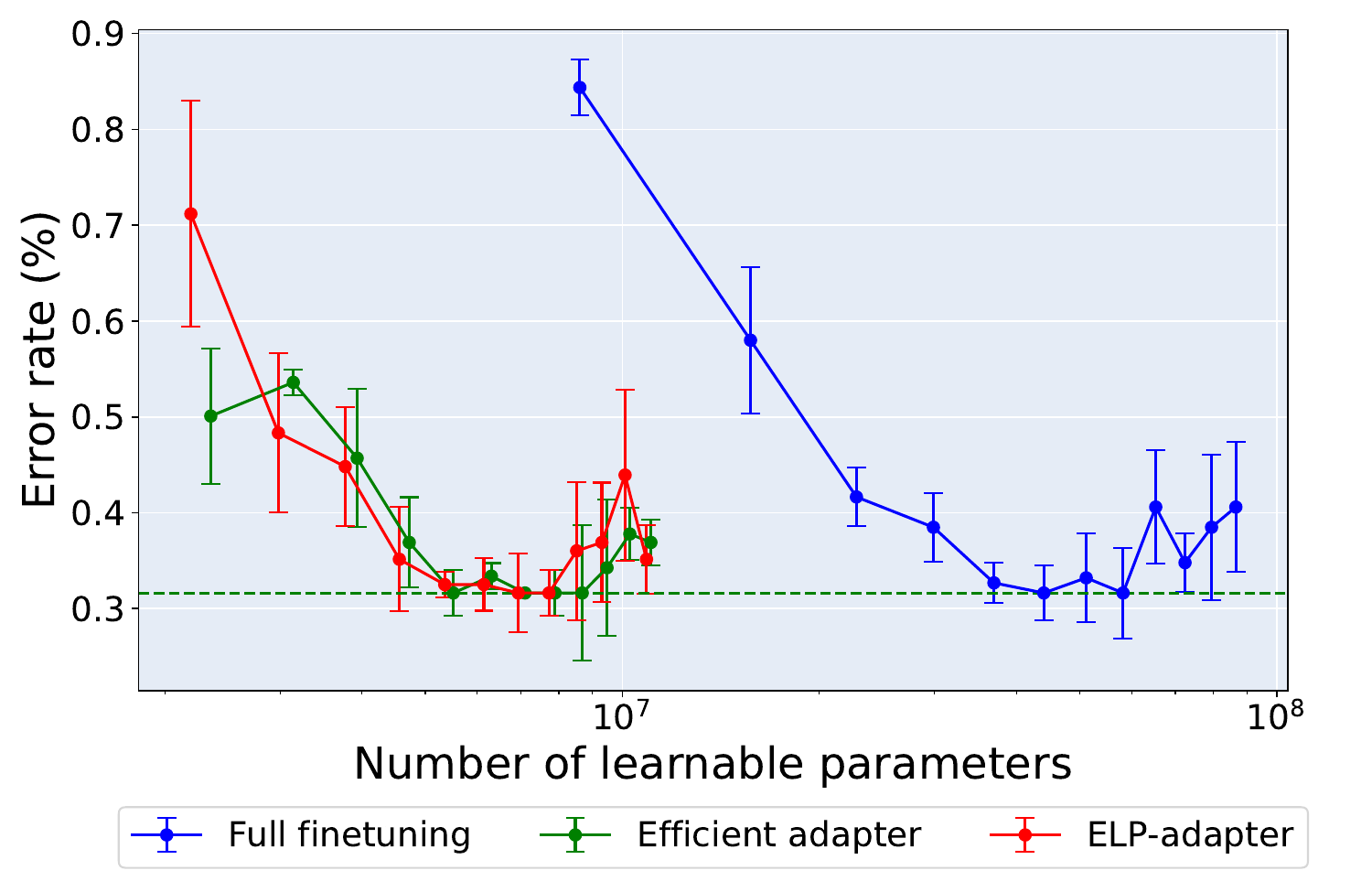}
\caption{Trade-off between number of learnable parameters and SIC performance in terms of ER. \wavlmisuseds}
\label{fig:tradeoff_sic}
\end{figure}

%% file: sec/9_analysis.tex
\section{Discussion and analysis}
\label{sec:analysis}

\subsection{Layer weight analysis}
To analyze the contribution of each layer, \cref{fig_weight_analysis} visualizes the learned
weight coefficients $w_{l}$ in Eqs.~(\ref{weight_tuning}) and (\ref{qe:bara_ladapter}) for each layer $l = 1, 2, \cdots, 12$ obtained in weight tuning, L-adapter tuning and ELP-adapter tuning.

In ASR, the weights obtained from the upper layers (layers from 9 to 12) tend to be larger.
With ELP-adapter tuning, the topmost layer has the largest weight for four models (b-e).
This indicates that the updates through E-adapters connected in series contribute significantly to the ASR performance.
\input{fig/fig_weight_analysis}

\input{fig/fig_ladapter}

In ASV, the weight distribution of weight tuning is similar to that of ASR.
In contrast, the distribution is shifted clearly with L-adapter tuning and ELP-adapter tuning, resulting in larger weights in the lower layers (layers 3 to 5).
This suggests that extracting features to identify speakers is not straightforward with the low-level features obtained from the frozen lower layers, but L-adapters provide a means to better leverage the lower layers.

In SER, all layers are leveraged almost equally.
This shows that the combination of features extracted from lower to higher layers contributes to identifying emotions.

In SIC, the weights of upper layers are relatively larger than those of lower layers with weight tuning.
However, all layers are leveraged almost equally in L-adapter tuning and ELP-adapter tuning.
While high-level features are crucial for SIC, this result indicates that this SIC task could be solvable even with a smaller number of layers.

\subsection{Adapter configurations}
\label{sec:padapterdetails}
\input{fig/fig_padapter}
\input{tab/tab_architecture_L_adapter}
\input{tab/tab_architecture_P_adapter}

\subsubsection{L-adapter configurations}
To investigate the most effective configuration for L-adapters, we conducted experiments with nine configurations in Table~\ref{tab:architecture_L_adapter}.
\cref{fig:ladapter} illustrates these configurations.
Configuration (A) uses only the weighted sum (``Weight'' in the table) of the encoder outputs, which has 12 parameters as described in \cref{weight_tuning}.
Configuration (B) introduces a single LayerNorm into each adapter, resulting in 18k learnable parameters.
The presence of LayerNorm potentially facilitates learning more effective representations by reducing internal covariate shift. The performance improvement in ASR, ASV, and SER suggests that this normalization step helps to extract more useful features from the speech signal for these tasks.
Configuration (C) adds the activation function to (B), but the performance of ASR and ER was degraded.
Applying the activation function was not effective because all encoder layers are frozen.

\input{tab/tab_limitation}

Configuration (D) makes all LayerNrom parameters learnable in the backbone self-supervised models. Note that this is what we called ``Weight tuning'' in previous sections.
As shown in the table, the performance on all tasks was improved when comparing (A) and (D).
Further, (E) incorporates a learnable fully connected layer into each L-adapter.
This significantly the perfomance of ASR, ASV and ER.
Although fully connected layers increase the number of learnable parameters to 4.75M, this is considered the minimum requirement.

Configurations (F), (G), and (H) add activation function, LayerNorm and both of them to (E), respectively.
As shown, (H) achieved the best or second best performance among all configurations on all tasks.
Finally, configuration (I) investigates the effectiveness of the skip connection, but we did not observe any significant performance improvement.
From these results, we conclude that (H), which is the default configuration of L-adapters, represents the optimal balance of efficiency and effectiveness.

\subsubsection{P-adapter configuration}
\cref{tab:architecture_P_adapter} compares P-adapter configurations described in \cref{sec:P-adapter}.
Configurations (A) and (B) use the prefix P-adapter and its nonlinear extension, respectively.
As shown, the nonlinear extension improved the performance on all tasks.
However, with the suffix P-adapter in (C) and (D), the nonlinear extension improved the performance only on ASV.
The default setting we used was (C), but these results suggest that the best configuration of P-adapter depends on the task.

\subsection{Limitations}
When a large amount of data is available for fine-tuning, it is advantageous to update more parameters.
Consequently, ELP-adapter tuning does not always outperform full fine-tuning.
To analyze this limitation,
we compared full fine-tuning and ELP-adapter tuning using larger training data for ASR and ASV with the WavLM+ backbone.

For ASR, the train-clean-100 set of LibriSpeech consisting of 100 hours of clean speech data was used for training, and the test-clean set was used for testing.
Results are reported in Table \ref{tab:limitation} with and without the 4-gram language model of LibriSpeech, applied in the same way as in~\cite{baevski2020wav2vec2}.
As shown, ELP-adapter tuning approaches the performance of full fine-tuning but falls short by 0.06 and 0.07 points in WER, with and without the language model, respectively.

For ASV, the VoxCeleb2 training set consisting of 1.1 million audio utterances from 5,994 speakers was used for training, and the VoxCeleb1 test set was used for testing.
Results are reported in Table~\ref{tab:limitation} in two settings: 1) the same setting as in Section V, where the linear head is used with the cross-entropy loss for 6 epochs, and
2) the x-vector setting~\cite{yang2024superb}, where the x-vector model~\cite{Snyder2018xvector} is used as a downstream head with the AMM softmax loss~\cite{Wang2018ammsoftmax} without noise-based augmentation for 12 epochs.
In contrast to ASR, ELP-adapter tuning outperformed full fine-tuning by 0.61 and 0.33 points, with the linear and x-vector heads, respectively.

%% file: fig/fig_weight_analysis.tex
\begin{figure*}
\centering
\includegraphics[width=\linewidth]{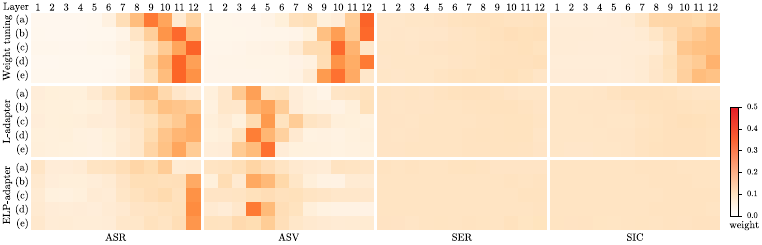}
\caption{Weight coefficients $w_{l}$ for layers $l = 1, 2, \cdots, 12$. Results of five self-supervised models are visualized: (a) Wav2vec2.0, (b) HuBERT, (c) ContentVec, (d) WavLM and (e) WavLM+.
}
\label{fig_weight_analysis}
\end{figure*}

%% file: fig/fig_ladapter.tex
\begin{figure}
\centering
\includegraphics[width=\linewidth]{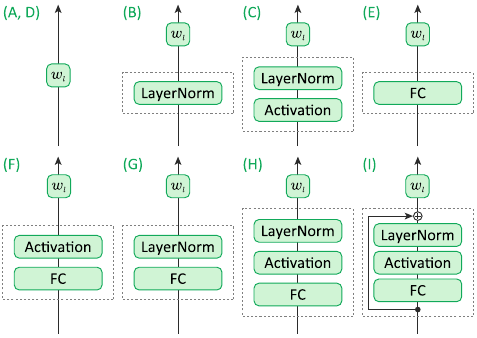}
\caption{Nine L-adapter configurations.
Configurations (A) to (I) are corresponding to those in \cref{tab:architecture_L_adapter}.}
\label{fig:ladapter}
\end{figure}

%% file: fig/fig_padapter.tex
\begin{figure}
\centering
\includegraphics[width=\linewidth]{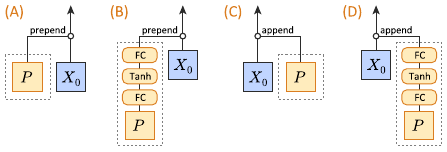}
\caption{Four P-adapter configurations.
$X_{0}$ is the output of the CNN encoder.
$P$ is a learnable matrix.
Configurations (A) to (D) are corresponding to those in \cref{tab:architecture_P_adapter}.}
\label{fig:padapter}
\end{figure}

%% file: tab/tab_architecture_L_adapter.tex
\begin{table*}[t]
\setlength{\tabcolsep}{5.5pt}
\centering
\caption{Evaluation of L-adapter variants.
The best and second best results are highlighted in bold and underlined, respectively. \confidenceshort}
\label{tab:architecture_L_adapter}
\begin{tabular}{c|cc|cccc|c|cccc}
\toprule
Config. & Weight & B.Norm & FC layer & Activation & LayerNorm & Skip & \# Params & ASR & ASV & ER & IC \\
\midrule
(A) & $\checkmark$ & & & &  & & 12 & $23.8_{\pm0.135}$ & $5.65_{\pm 0.258}$ & $28.9_{\pm 0.522}$ & $0.73_{\pm 0.086}$ \\
(B) & $\checkmark$ & & & & $\checkmark$ & & 0.018M & $21.8_{\pm0.173}$ & $4.76_{\pm 0.075}$ & $25.8_{\pm 0.442}$ & $0.89_{\pm 0.084}$ \\
(C) & $\checkmark$ & & & $\checkmark$ & $\checkmark$ & & 0.018M & $21.7_{\pm 0.297}$ & $5.25_{\pm 0.234}$ & $26.9_{\pm 0.443}$ & $0.73_{\pm0.089}$ \\
\midrule
(D) & $\checkmark$ & $\checkmark$ & & &  & & 0.037M & $21.5_{\pm 0.387}$ & $5.23_{\pm 0.062}$ & $28.4_{\pm 0.571}$ & $0.45_{\pm 0.101}$\\
(E) & $\checkmark$ & $\checkmark$ & $\checkmark$ & & & & 4.75M & $16.8_{\pm 0.053}$ & $3.71_{\pm 0.077}$ & $24.2_{\pm 0.312}$ & $0.73_{\pm 0.114}$ \\
(F) & $\checkmark$ & $\checkmark$ & $\checkmark$ & $\checkmark$ & & & 4.75M & $\underline{10.2}_{\pm 0.058}$ & $4.22_{\pm 0.098}$ & ${22.8}_{\pm 0.553}$ & $0.67_{\pm 0.038}$ \\
(G) & $\checkmark$ & $\checkmark$ & $\checkmark$ & & $\checkmark$ & & 4.77M & $13.5_{\pm 0.136}$ & ${\bm{2.73}}_{\pm 0.058}$ & $23.3_{\pm 0.511}$ & ${\bm{0.32}}_{\pm 0.038}$ \\
(H) & $\checkmark$ & $\checkmark$ & $\checkmark$ & $\checkmark$ & $\checkmark$ & & 4.77M &  ${\bm{9.50}}_{\pm 0.089}$ & $\underline{2.74}_{\pm 0.097}$ & $\underline{21.1}_{\pm 0.522}$ & $\underline{0.33}_{\pm 0.014}$ \\
(I) & $\checkmark$ & $\checkmark$ & $\checkmark$ & $\checkmark$ & $\checkmark$ & $\checkmark$ & 4.77M & ${10.4}_{\pm0.082}$ & ${2.78}_{\pm 0.062}$ & ${\bm{20.0}}_{\pm 0.281}$ & ${0.42}_{\pm 0.090}$\\
\bottomrule
\end{tabular}
\vspace{4pt}
\end{table*}

%% file: tab/tab_architecture_P_adapter.tex
\begin{table*}[t]
\centering
\caption{Evaluation of P-adapter variants. The best and second best results are highlighted in bold and underlined, respectively. \confidenceshort}
\label{tab:architecture_P_adapter}
\begin{tabular}{l|ccc|c|cccc}
\toprule
& Prefix & Suffix & Nonlinear & \# Params & ASR & ASV & ER & IC \\
\midrule
(A) & $\checkmark$ & & & 3,840  & $23.78_{\pm 0.147}$ & 	$5.52_{\pm 0.295}$ & 	$30.81_{\pm 0.728}$ & 	$0.89_{\pm 0.409}$ \\
(B) & $\checkmark$ &  & $\checkmark$ & 1.19M  & $\bm{23.27}_{\pm 0.247}$ & 	${5.46}_{\pm 0.401}$ & 	${30.52}_{\pm 0.603}$ & 	${0.87}_{\pm 0.285}$ \\
\midrule
(C) & & $\checkmark$ & & 3,840  & ${23.68}_{\pm 0.297}$ & $5.62_{\pm 0.114}$ & 	$\bf{30.14}_{\pm 1.106}$ & 	$\bf{0.80}_{\pm 0.300}$ \\
(D) & & $\checkmark$ & $\checkmark$ & 1.19M &  ${23.69}_{\pm 0.624}$ & 	$\bm{5.25}_{\pm 0.216}$ & 	$31.39_{\pm 0.625}$ & 	$1.00_{\pm 0.115}$ \\
\bottomrule
\end{tabular}
\end{table*}

%% file: tab/tab_limitation.tex
\def\asrelp{4.60}
\def\asrelpwlm{3.02}
\def\asrfull{4.54}
\def\asrfullwlm{2.95}
\def\asvelp{2.06}
\def\asvelpwxvec{1.13}
\def\asvfull{2.67}
\def\asvfullwxvec{1.46}

\begin{table}[t]
\setlength{\tabcolsep}{5pt}
\centering
\caption{
Evaluation using larger amounts of training data.
}
\begin{tabular}{l|cc|cc}
\toprule
\multirow{2}{*}{Fine-tuning method} & \multicolumn{2}{c|}{ASR} & \multicolumn{2}{c}{ASV}\\
& w/o LM & w/ LM & w/ linear head & w/ x-vector\\
\midrule
ELP-adapter tuning & \asrelp & \asrelpwlm & \textbf{\asvelp} & \textbf{\asvelpwxvec} \\
Full fine-tuning & \textbf{\asrfull} & \textbf{\asrfullwlm} & \asvfull & \asvfullwxvec\\
\bottomrule
\end{tabular}
\label{tab:limitation}
\end{table}

%% file: sec/10_conclusion.tex
\section{Conclusion}
\label{sec:conclusion}

This paper proposed ELP-adapter tuning, a novel method for parameter-efficient fine-tuning for various speech processing tasks.
We introduced three types of adapter, namely
E-adapters for learning fine-grained speech representation,
L-adapters for extracting non-linguistic features from lower layers, and P-adapters for improving training efficiency with pseudo features.
The results of experiments on ASR, ASV, SER and SIC tasks demonstrated the effectiveness and efficiency of the proposed method compared to conventional fine-tuning methods.
Future work will focus on further improving efficiency through automatic pruning of adapter types and neural architecture search, as well as applying adapters to more complex and generative tasks such as spoken question answering~\cite{sqa1, sqa2, sqa3, sqa4, sqa5, sqa6} and voice conversion~\cite{vc0, vc1, vc2, vc3, vc4}.
Multi-modal adapters for speaker verification and emotion recognition over both audio and visual streams will also be investigated.

\section*{Acknowledgements}
This work was supported by JSPS KAKENHI Grant Numbers 22K12089 and 23H00490.

%% file: main.bbl
\begin{thebibliography}{10}
\providecommand{\url}[1]{#1}
\csname url@samestyle\endcsname
\providecommand{\newblock}{\relax}
\providecommand{\bibinfo}[2]{#2}
\providecommand{\BIBentrySTDinterwordspacing}{\spaceskip=0pt\relax}
\providecommand{\BIBentryALTinterwordstretchfactor}{4}
\providecommand{\BIBentryALTinterwordspacing}{\spaceskip=\fontdimen2\font plus
\BIBentryALTinterwordstretchfactor\fontdimen3\font minus \fontdimen4\font\relax}
\providecommand{\BIBforeignlanguage}[2]{{%
\expandafter\ifx\csname l@#1\endcsname\relax
\typeout{** WARNING: IEEEtran.bst: No hyphenation pattern has been}%
\typeout{** loaded for the language `#1'. Using the pattern for}%
\typeout{** the default language instead.}%
\else
\language=\csname l@#1\endcsname
\fi
#2}}
\providecommand{\BIBdecl}{\relax}
\BIBdecl

\bibitem{schneider2020wav2vec}
S.~Schneider, A.~Baevski, R.~Collobert, and M.~Auli, ``wav2vec: Unsupervised pre-training for speech recognition,'' in \emph{Proc. Interspeech}, 2019.

\bibitem{baevski2020wav2vec2}
A.~Baevski, H.~Zhou, A.~Mohamed, and M.~Auli, ``wav2vec 2.0: A framework for self-supervised learning of speech representations,'' in \emph{Proc. Annual Conference on Neural Information Processing Systems (NeurIPS)}, 2020.

\bibitem{wang2021unispeech}
C.~Wang, Y.~Wu, Y.~Qian, K.~Kumatani, S.~Liu, F.~Wei, M.~Zeng, and X.~Huang, ``Unispeech: Unified speech representation learning with labeled and unlabeled data,'' in \emph{Proc. International Conference on Machine Learning (ICML)}, 2021.

\bibitem{hsu2021hubert}
W.-N. Hsu, B.~Bolte, Y.-H.~H. Tsai, K.~Lakhotia, R.~Salakhutdinov, and A.~Mohamed, ``{HuBERT}: Self-supervised speech representation learning by masked prediction of hidden units,'' \emph{IEEE/ACM Transactions on Audio, Speech, and Language Processing (TASLP)}, vol.~29, pp. 3451--3460, 2021.

\bibitem{chen2022unispeech-sat}
S.~Chen, Y.~Wu, C.~Wang, Z.~Chen, Z.~Chen, S.~Liu, J.~Wu, Y.~Qian, F.~Wei, J.~Li, and X.~Yu, ``Unispeech-sat: Universal speech representation learning with speaker aware pre-training,'' in \emph{Proc. IEEE International Conference on Acoustics, Speech and Signal Processing (ICASSP)}, 2022.

\bibitem{chen2022wavlm}
S.~Chen, C.~Wang, Z.~Chen, Y.~Wu, S.~Liu, Z.~Chen, J.~Li, N.~Kanda, T.~Yoshioka, X.~Xiao \emph{et~al.}, ``{WavLM}: Large-scale self-supervised pre-training for full stack speech processing,'' \emph{IEEE Journal of Selected Topics in Signal Processing (JSTSP)}, 2022.

\bibitem{qian2022contentvec}
K.~Qian, Y.~Zhang, H.~Gao, J.~Ni, C.-I. Lai, D.~Cox, M.~Hasegawa-Johnson, and S.~Chang, ``{ContentVec}: An improved self-supervised speech representation by disentangling speakers,'' in \emph{Proc. International Conference on Machine Learning (ICML)}, 2022.

\bibitem{chen2020simclr}
T.~Chen, S.~Kornblith, M.~Norouzi, and G.~Hinton, ``A simple framework for contrastive learning of visual representations,'' in \emph{Proc. International Conference on Machine Learning (ICML)}, 2020.

\bibitem{jiang2021speechsimclr}
D.~Jiang, W.~Li, M.~Cao, W.~Zou, and X.~Li, ``Speech {SimCLR}: Combining contrastive and reconstruction objective for self-supervised speech representation learning,'' in \emph{Proc. Interspeech}, 2021.

\bibitem{huh2020augmentation}
J.~Huh, H.~S. Heo, J.~Kang, S.~Watanabe, and J.~S. Chung, ``Augmentation adversarial training for unsupervised speaker recognition,'' in \emph{Proc. NeurIPS Workshop on Self-Supervised Learning for Speech and Audio Processing}, 2020.

\bibitem{inoue2020semi}
N.~Inoue and K.~Goto, ``Semi-supervised contrastive learning with generalized contrastive loss and its application to speaker recognition,'' in \emph{Proc. Asia-Pacific Signal and Information Processing Association Annual Conference and Summit (APSIPA ASC)}, 2020.

\bibitem{finetune-sv}
N.~Vaessen and D.~A. Van~Leeuwen, ``Fine-tuning wav2vec2 for speaker recognition,'' in \emph{Proc. IEEE International Conference on Acoustics, Speech and Signal Processing (ICASSP)}, 2022, pp. 7967--7971.

\bibitem{fan21wav2vecsv}
Z.~Fan, M.~Li, S.~Zhou, and B.~Xu, ``Exploring wav2vec 2.0 on speaker verification and language identification,'' in \emph{Proc. Interspeech}, 2021, pp. 1509--1513.

\bibitem{lee22wav2vecsv}
J.~W. Lee, E.~Kim, J.~Koo, and K.~Lee, ``Representation selective self-distillation and wav2vec 2.0 feature exploration for spoof-aware speaker verification,'' in \emph{Proc. Interspeech}, 2022, pp. 2898--2902.

\bibitem{peng2023improving}
J.~Peng, O.~Plchot, T.~Stafylakis, L.~Mo{\v{s}}ner, L.~Burget, and J.~{\v{C}}ernock{\'y}, ``Improving speaker verification with self-pretrained transformer models,'' in \emph{Proc. Interspeech}, 2023.

\bibitem{finetune-er}
L.~Pepino, P.~Riera, and L.~Ferrer, ``Emotion recognition from speech using wav2vec 2.0 embeddings,'' in \emph{Proc. Interspeech}, 2021, pp. 3400--3404.

\bibitem{pepino21wav2vecer}
------, ``Emotion recognition from speech using wav2vec 2.0 embeddings,'' in \emph{Proc. Interspeech}, 2021, pp. 3400--3404.

\bibitem{sqa1}
C.~You, N.~Chen, and Y.~Zou, ``Self-supervised contrastive cross-modality representation learning for spoken question answering,'' in \emph{Findings of Empirical Methods in Natural Language Processing (EMNLP Findings)}, 2021, pp. 28--39.

\bibitem{sqa2}
------, ``Knowledge distillation for improved accuracy in spoken question answering,'' in \emph{Proc. IEEE International Conference on Acoustics, Speech and Signal Processing (ICASSP)}, 2021.

\bibitem{sqa3}
------, ``Contextualized attention-based knowledge transfer for spoken conversational question answering,'' in \emph{Proc. Interspeech}, 2021, pp. 3211--3215.

\bibitem{sqa4}
C.~You, N.~Chen, F.~Liu, S.~Ge, X.~Wu, and Y.~Zou, ``End-to-end spoken conversational question answering: Task, dataset and model,'' in \emph{Findings of Annual Conference of the North American Chapter of the Association for Computational Linguistics (NAACL)}, 2022, pp. 3211--3215.

\bibitem{sqa5}
N.~Chen, C.~You, and Y.~Zou, ``Self-supervised dialogue learning for spoken conversational question answering,'' in \emph{Proc. Interspeech}, 2021, pp. 231--235.

\bibitem{sqa6}
C.~You, N.~Chen, and Y.~Zou, ``Mrd-net: Multi-modal residual knowledge distillation for spoken question answering,'' in \emph{Proc. International Joint Conference on Artificial Intelligence (IJCAI)}, 2021, pp. 3985--3991.

\bibitem{vc0}
H.~Huang, L.~Wang, J.~Yang, Y.~Hu, and L.~He, ``{W2VC: WavLM representation based one-shot voice conversion with gradient reversal distillation and CTC supervision},'' \emph{{EURASIP Journal on Audio, Speech, and Music Processing}}, vol. 2023, no.~1, p.~45, 2023.

\bibitem{vc1}
J.~hao Lin, Y.~Y. Lin, C.-M. Chien, and H.~yi~Lee, ``{S2VC}: A framework for any-to-any voice conversion with self-supervised pretrained representations,'' in \emph{Proc. Interspeech}, 2021, pp. 836--840.

\bibitem{vc2}
M.~Baas, B.~van Niekerk, and H.~Kamper, ``Voice conversion with just nearest neighbors,'' in \emph{Proc. Interspeech}, 2023.

\bibitem{vc3}
J.~Lim and K.~Kim, ``Wav2vec-vc: Voice conversion via hidden representations of wav2vec 2.0,'' in \emph{Proc. IEEE International Conference on Acoustics, Speech and Signal Processing (ICASSP)}, 2024, pp. 10\,326--10\,330.

\bibitem{vc4}
H.-S. Tsai, H.-J. Chang, W.-C. Huang, Z.~Huang, K.~Lakhotia, S.~wen Yang, S.~Dong, A.~T. Liu, C.-I. Lai, J.~Shi, X.~Chang, P.~Hall, H.-J. Chen, S.-W. Li, S.~Watanabe, A.~rahman Mohamed, and H.~yi~Lee, ``{SUPERB-SG}: Enhanced speech processing universal performance benchmark for semantic and generative capabilities,'' in \emph{Proc. Annual Meeting of the Association for Computational Linguistics (ACL)}, vol. abs/2203.06849, 2022, pp. 836--840.

\bibitem{nlp-adapter}
N.~Houlsby, A.~Giurgiu, S.~Jastrzebski, B.~Morrone, Q.~De~Laroussilhe, A.~Gesmundo, M.~Attariyan, and S.~Gelly, ``{Parameter-Efficient Transfer Learning for NLP},'' in \emph{Proc. International Conference on Machine Learning (ICML)}, 2019, pp. 2790--2799.

\bibitem{bert}
J.~Devlin, M.-W. Chang, K.~Lee, and K.~Toutanova, ``{BERT}: Pre-training of deep bidirectional transformers for language understanding,'' in \emph{Proc. Annual Conference of the North American Chapter of the Association for Computational Linguistics (NAACL)}, 2019.

\bibitem{lin2020exploring}
Z.~Lin, A.~Madotto, and P.~Fung, ``Exploring versatile generative language model via parameter-efficient transfer learning,'' in \emph{Findings of Empirical Methods in Natural Language Processing (EMNLP Findings)}, 2020.

\bibitem{Guo2021AdaptiveAdapters}
J.~Guo, Z.~Zhang, L.~Xu, X.~Chen, and E.~Chen, ``Adaptive adapters: An efficient way to incorporate {BERT} into neural machine translation,'' \emph{IEEE/ACM Transactions on Audio, Speech, and Language Processing (TASLP)}, vol.~29, pp. 1740--1751, 2021.

\bibitem{Zhang2023PoE}
C.~Zhang, L.~F. D'Haro, Q.~Zhang, and T.~Friedrichs, ``Poe: A panel of experts for generalized automatic dialogue assessment,'' \emph{IEEE/ACM Transactions on Audio, Speech, and Language Processing (TASLP)}, vol.~31, pp. 1234--1250, 2023.

\bibitem{rnnt-adapter}
A.~Kannan, A.~Datta, T.~N. Sainath, E.~Weinstein, B.~Ramabhadran, Y.~Wu, A.~Bapna, Z.~Chen, and S.~Lee, ``Large-scale multilingual speech recognition with a streaming end-to-end model,'' in \emph{Proc. Interspeech}, 2019.

\bibitem{hou21meta}
W.~Hou, Y.~Wang, S.~Gao, and T.~Shinozaki, ``Meta-adapter: Efficient cross-lingual adaptation with meta-learning,'' in \emph{Proc. IEEE International Conference on Acoustics, Speech and Signal Processing (ICASSP)}, 2021.

\bibitem{hou22adapters}
W.~Hou, H.~Zhu, Y.~Wang, J.~Wang, T.~Qin, R.~Xu, and T.~Shinozaki, ``Exploiting adapters for cross-lingual low-resource speech recognition,'' \emph{IEEE/ACM Transactions on Audio, Speech, and Language Processing (TASLP)}, vol.~30, pp. 317--329, 2022.

\bibitem{ctcattn-adapter}
G.~I. Winata, G.~Wang, C.~Xiong, and S.~Hoi, ``{Adapt-and-Adjust}: Overcoming the long-tail problem of multilingual speech recognition,'' in \emph{Proc. Interspeech}, 2021.

\bibitem{Qian2022LayerWiseFastAdaptation}
Y.~Qian, X.~Gong, and H.~Huang, ``Layer-wise fast adaptation for end-to-end multi-accent speech recognition,'' \emph{IEEE/ACM Transactions on Audio, Speech, and Language Processing (TASLP)}, vol.~30, pp. 2842--2853, 2022.

\bibitem{sl-adapter}
H.~Le, J.~Pino, C.~Wang, J.~Gu, D.~Schwab, and L.~Besacier, ``Lightweight adapter tuning for multilingual speech translation,'' in \emph{Proc. Annual Meeting of the Association for Computational Linguistics (ACL)}, 2021.

\bibitem{speech-adapter}
B.~Thomas, S.~Kessler, and S.~Karout, ``Efficient adapter transfer of self-supervised speech models for automatic speech recognition,'' in \emph{Proc. IEEE International Conference on Acoustics, Speech and Signal Processing (ICASSP)}, 2022, pp. 7102--7106.

\bibitem{Chen2022EfficientTuning}
Z.-C. Chen, C.-L. Fu, C.-Y. Liu, S.-W. Li, and H.-Y. Lee, ``Exploring efficient-tuning methods in self-supervised speech models,'' in \emph{Proc. IEEE Workshop on Spoken Language Technology (SLT)}, 2022.

\bibitem{hu22lora}
E.~J. Hu, Y.~Shen, P.~Wallis, Z.~Allen-Zhu, Y.~Li, S.~Wang, L.~Wang, and W.~Chen, ``{LoRA}: Low-rank adaptation of large language models,'' in \emph{Proc. International Conference on Learning Representations (ICLR)}, 2022.

\bibitem{analysis1}
A.~Pasad, J.-C. Chou, and K.~Livescu, ``Layer-wise analysis of a self-supervised speech representation model,'' in \emph{Proc. IEEE Workshop on Automatic Speech Recognition and Understanding (ASRU)}, 2021, pp. 914--921.

\bibitem{analysis2}
J.~Shah, Y.~K. Singla, C.~Chen, and R.~R. Shah, ``What all do audio transformer models hear? probing acoustic representations for language delivery and its structure,'' \emph{arXiv preprint arXiv:2101.00387}, 2021.

\bibitem{otake2023parameter}
S.~Otake, R.~Kawakami, and N.~Inoue, ``Parameter efficient transfer learning for various speech processing tasks,'' in \emph{Proc. IEEE International Conference on Acoustics, Speech and Signal Processing (ICASSP)}, 2023.

\bibitem{li21prefixtuning}
X.~L. Li and P.~Liang, ``Prefix-tuning: Optimizing continuous prompts for generation,'' in \emph{Proc. Joint Conference of Annual Meeting of the Association for Computational Linguistics and International Joint Conference on Natural Language Processing (ACL-IJCNLP)}, 2021, pp. 4582--4597.

\bibitem{ba16layernorm}
J.~L. Ba, J.~R. Kiros, and G.~E. Hinton, ``Layer normalization,'' \emph{arXiv preprint arXiv:1607.06450}, 2016.

\bibitem{Hendrycks16GELU}
D.~Hendrycks and K.~Gimpel, ``Gaussian error linear units (gelus),'' \emph{arXiv preprint arXiv:1606.08415}, 2016.

\bibitem{panayotov2015librispeech}
V.~Panayotov, G.~Chen, D.~Povey, and S.~Khudanpur, ``Librispeech: An asr corpus based on public domain audio books,'' in \emph{Proc. IEEE International Conference on Acoustics, Speech and Signal Processing (ICASSP)}, 2015, pp. 5206--5210.

\bibitem{chi2022xlm}
Z.~Chi, S.~Huang, L.~Dong, S.~Ma, B.~Zheng, S.~Singhal, P.~Bajaj, X.~Song, X.-L. Mao, H.~Huang, and F.~Wei, ``{XLM}-{E}: Cross-lingual language model pre-training via {ELECTRA},'' in \emph{Proc. Annual Meeting of the Association for Computational Linguistics (ACL)}, 2022, pp. 6170--6182.

\bibitem{chen21gigaspeech}
G.~Chen, S.~Chai, G.-B. Wang, J.~Du, W.~Zhang, C.~Weng, D.~Su, D.~Povey, J.~Trmal, J.~Zhang, M.~Jin, S.~Khudanpur, S.~Watanabe, S.~Zhao, W.~Zou, X.~Li, X.~Yao, Y.~Wang, Z.~You, and Z.~Yan, ``Gigaspeech: An evolving, multi-domain asr corpus with 10,000 hours of transcribed audio,'' in \emph{Proc. Interspeech}, 2021.

\bibitem{wang21voxpopuli}
C.~Wang, M.~Riviere, A.~Lee, A.~Wu, C.~Talnikar, D.~Haziza, M.~Williamson, J.~Pino, and E.~Dupoux, ``{V}ox{P}opuli: A large-scale multilingual speech corpus for representation learning, semi-supervised learning and interpretation,'' in \emph{Proc. Joint Conference of Annual Meeting of the Association for Computational Linguistics and International Joint Conference on Natural Language Processing (ACL-IJCNLP)}, 2021, pp. 993--1003.

\bibitem{graves2006ctc}
A.~Graves, S.~Fernandez, F.~Gomez, and J.~Schmidhuber, ``Connectionist temporal classification: labelling unsegmented sequence data with recurrent neural networks,'' in \emph{Proc. International Conference on Machine Learning (ICML)}, 2006.

\bibitem{superb}
S.-w. Yang, P.-H. Chi, Y.-S. Chuang, C.-I.~J. Lai, K.~Lakhotia, Y.~Y. Lin, A.~T. Liu, J.~Shi, X.~Chang, G.-T. Lin, T.-H. Huang, W.-C. Tseng, K.-t. Lee, D.-R. Liu, Z.~Huang, S.~Dong, S.-W. Li, S.~Watanabe, A.~Mohamed, and H.-y. Lee, ``{SUPERB}: Speech processing universal performance benchmark,'' in \emph{Proc. Interspeech}, 2021, pp. 1194--1198.

\bibitem{librilight}
J.~Kahn, M.~Rivi{\`e}re, W.~Zheng, E.~Kharitonov, Q.~Xu, P.-E. Mazar{\'e}, J.~Karadayi, V.~Liptchinsky, R.~Collobert, C.~Fuegen \emph{et~al.}, ``{Libri-Light}: A benchmark for asr with limited or no supervision,'' in \emph{Proc. IEEE International Conference on Acoustics, Speech and Signal Processing (ICASSP)}, 2020, pp. 7669--7673.

\bibitem{voxceleb}
A.~Nagrani, J.~S. Chung, and A.~Zisserman, ``{VoxCeleb}: A large-scale speaker identification dataset,'' in \emph{Proc. Interspeech}, 2017, pp. 2616--2620.

\bibitem{snorm1}
Z.~N. Karam, W.~M. Campbell, and N.~Dehak, ``Towards reduced false-alarms using cohorts,'' in \emph{Proc. IEEE International Conference on Acoustics, Speech and Signal Processing (ICASSP)}, 2011, pp. 4512--4515.

\bibitem{snorm2}
S.~Cumani, P.~Batzu, D.~Colibro, C.~Vair, P.~Laface, and V.~Vasilakakis, ``Comparison of speaker recognition approaches for real applications,'' in \emph{Proc. Interspeech}, 2011, pp. 2365--2368.

\bibitem{iemocap}
C.~Busso, M.~Bulut, C.~Lee, A.~Kazemzadeh, E.~Mower, J.~C. S.~Kim, S.~Lee, and S.~Narayanan, ``{IEMOCAP}: Interactive emotional dyadic motion capture database,'' \emph{Language resources and evaluation}, vol.~42, no.~4, pp. 335--359, 2008.

\bibitem{erpaper}
H.~M. Fayek, M.~Lech, and L.~Cavedon, ``Evaluating deep learning architectures for speech emotion recognition,'' \emph{Neural Networks}, vol.~92, pp. 60--68, 2017.

\bibitem{speechcommand}
L.~Lugosch, M.~Ravanelli, P.~Ignoto, V.~S. Tomar, and Y.~Bengio, ``Speech model pre-training for end-to-end spoken language understanding,'' in \emph{Proc. Interspeech}, 2019.

\bibitem{yang2024superb}
S.-w. Yang, H.-J. Chang, Z.~Huang, A.~T. Liu, C.-I. Lai, H.~Wu, J.~Shi, X.~Chang, H.-S. Tsai, W.-C. Huang \emph{et~al.}, ``A large-scale evaluation of speech foundation models,'' \emph{IEEE/ACM Transactions on Audio, Speech, and Language Processing (TASLP)}, 2024.

\bibitem{Snyder2018xvector}
D.~Snyder, D.~Garcia-Romero, G.~Sell, D.~Povey, and S.~Khudanpur, ``X-vectors: Robust dnn embeddings for speaker recognition,'' in \emph{Proc. IEEE International Conference on Acoustics, Speech and Signal Processing (ICASSP)}, 2018, pp. 5329--5333.

\bibitem{Wang2018ammsoftmax}
F.~Wang, J.~Cheng, W.~Liu, and H.~Liu, ``Additive margin softmax for face verification,'' \emph{IEEE Signal Processing Letters}, vol.~25, no.~7, pp. 926--930, 2018.

\end{thebibliography}
